\documentclass[journal]{IEEEtran}

\usepackage{amsmath,amsfonts}
\usepackage{algorithmic}
\usepackage{array}
\usepackage[caption=false,font=normalsize,labelfont=sf,textfont=sf]{subfig}
\usepackage{textcomp}
\usepackage{stfloats}
\usepackage{url}
\usepackage{cite}
\usepackage{verbatim}
\usepackage{graphicx}
\usepackage{epstopdf}
\usepackage{caption}

\usepackage{subfig} 
\usepackage{caption}

\usepackage{multicol}
\usepackage{multirow}

\usepackage[british]{babel}

\usepackage[autostyle]{csquotes}

\usepackage[ruled,linesnumbered]{algorithm2e}
\usepackage{amsmath}

\def\BibTeX{{\rm B\kern-.05em{\sc i\kern-.025em b}\kern-.08em
    T\kern-.1667em\lower.7ex\hbox{E}\kern-.125emX}}
\usepackage{balance}
\begin{document}

\title{FedDistill: Global Model Distillation for Local Model De-Biasing in Non-IID Federated Learning}
\author{Changlin Song$^{1}$, Divya Saxena$^{1}$, Jiannong Cao, Yuqing Zhao$^{\ast}$
\thanks{$^{\ast}$corresponding author.}
\thanks{$^{1}$equal contribution.}
\\
charlemagnescl@outlook.com, \{divsaxen, csjcao, csyzhao1\}@comp.polyu.edu.hk
\\
Department of Computing, The Hong Kong Polytechnic University
}

\markboth{Journal of \LaTeX\ Class Files,~Vol.~18, No.~9, January~2024}%
{FedDistill: Global Model Distillation for Local Model De-biasing in Non-IID Federated Learning}

\maketitle

\begin{abstract}
Federated Learning (FL) is a novel approach that allows for collaborative machine learning while preserving data privacy by leveraging models trained on decentralized devices. However, FL faces challenges due to non-uniformly distributed (non-iid) data across clients, which impacts model performance and its generalization capabilities. To tackle the non-iid issue, recent efforts have utilized the global model as a teaching mechanism for local models. However, our pilot study shows that their effectiveness is constrained by imbalanced data distribution, which induces biases in local models and leads to a \textquote{local forgetting} phenomenon, where the ability of models to generalize degrades over time, particularly for underrepresented classes. This paper introduces FedDistill, a framework enhancing the knowledge transfer from the global model to local models, focusing on the issue of imbalanced class distribution. Specifically, FedDistill employs group distillation, segmenting classes based on their frequency in local datasets to facilitate a focused distillation process to classes with fewer samples. Additionally, FedDistill dissects the global model into a feature extractor and a classifier. This separation empowers local models with more generalized data representation capabilities and ensures more accurate classification across all classes. FedDistill mitigates the adverse effects of data imbalance, ensuring that local models do not forget underrepresented classes but instead become more adept at recognizing and classifying them accurately. Our comprehensive experiments demonstrate FedDistill's effectiveness, surpassing existing baselines in accuracy and convergence speed across several benchmark datasets.

\end{abstract}

\begin{IEEEkeywords}
Federated Learning, Non-iid, Knowledge Distillation
\end{IEEEkeywords}

\section{Introduction}

\IEEEPARstart{I}n the era of data-driven decision-making, Federated Learning (FL) has emerged as a pivotal technology, enabling collaborative model training across multiple devices while preserving data privacy. This approach not only enhances user privacy but also facilitates the utilization of decentralized data sources, making it a cornerstone for the next generation of machine learning applications, such as medical image processing \cite{medical2}, recommendation systems \cite{recommendation2}, and edge computing \cite{edge2}.


Despite its potential, FL faces critical challenges in achieving uniform model performance and generalization, particularly when dealing with non-independent and identically distributed (non-iid) data across different clients. This heterogeneity in data distribution leads to significant disparities in local model performance, as the local objective functions diverge from the global objective, pushing the aggregated global model away from the optimal solution. The non-iid problem is not only theoretical but exists in many real-world applications, potentially compromising the efficacy of FL systems.

\begin{figure}[!t]
\centering
\includegraphics[width=3.5in]{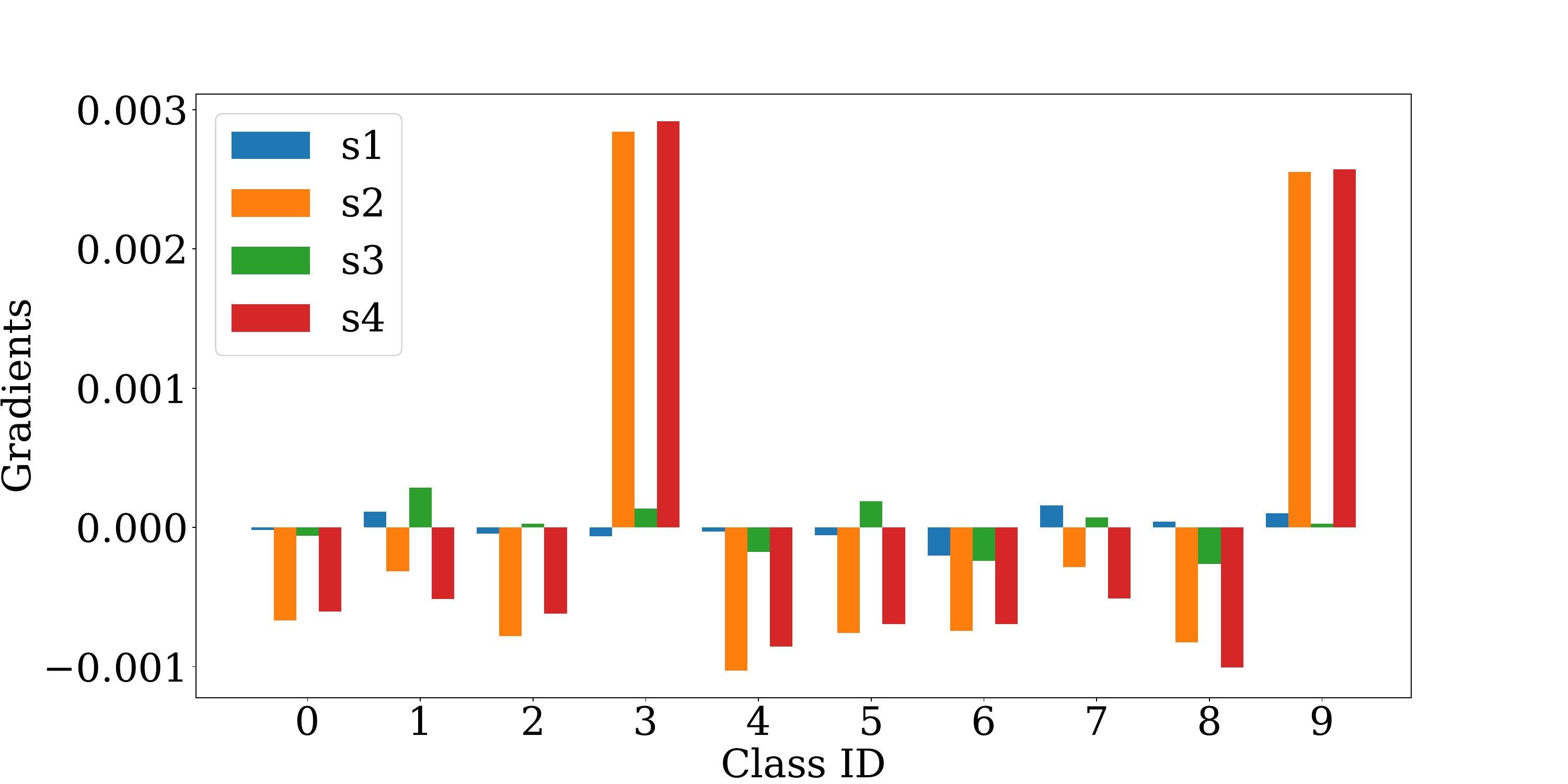}
\caption{Impact of imbalanced class distribution on a model's gradient updates. Gradient changes in a neural network across four scenarios (s1, s2, s3, s4) for ten classes (Class ID 0-9). The bars indicate gradient magnitudes, with colors representing different scenarios. Significant gradient increases for classes 3 and 9 in s2 and s4 point to shifts in learning focus due to class imbalance.}
\label{fig:Imbalanced_gradient}
\end{figure}



\begin{figure}[!t]
\centering
\subfloat[]{\includegraphics[width=1.7in]{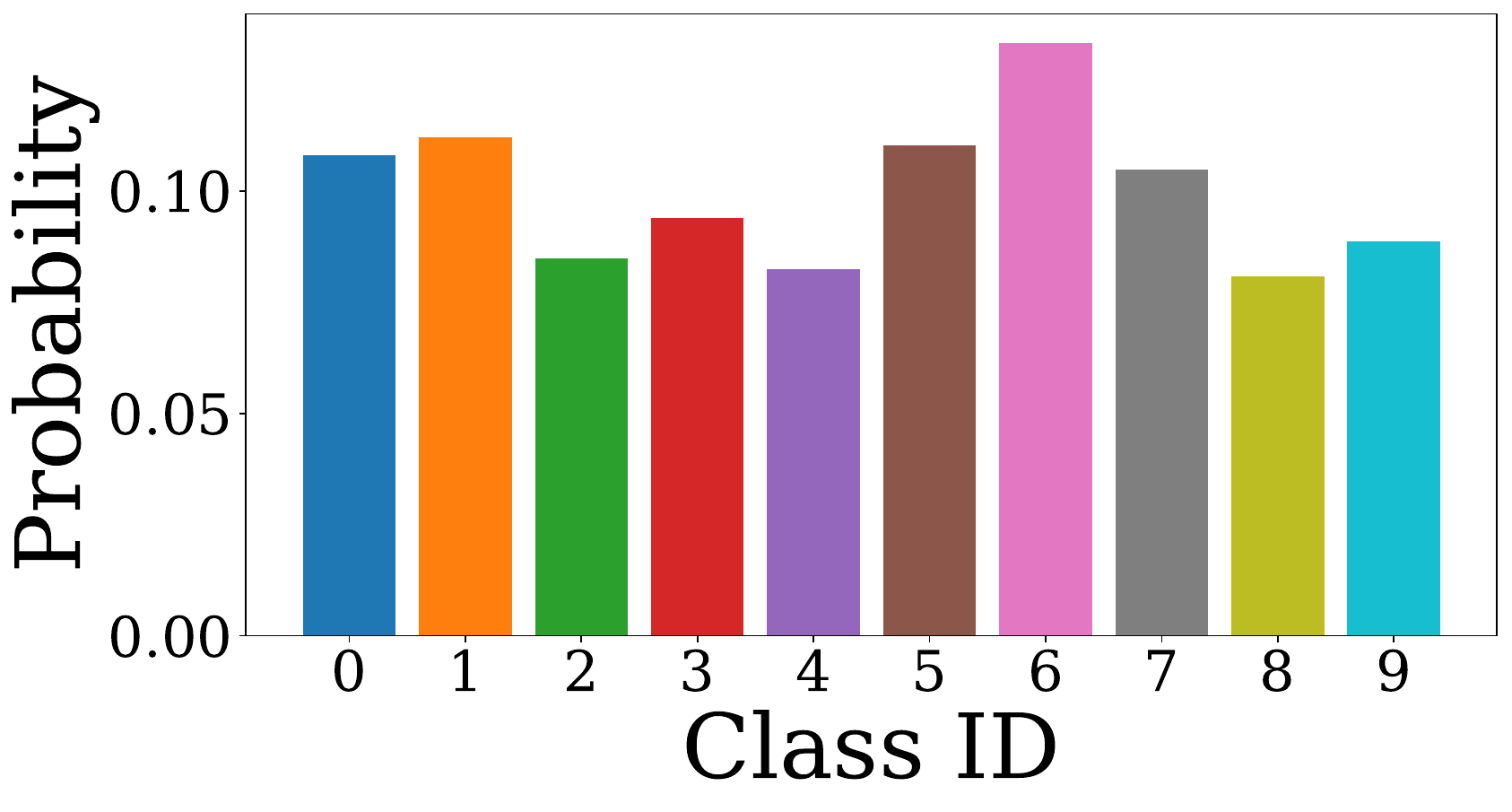}%
\label{fig:Imbalanced_pred_a}}
\hfil
\subfloat[]{\includegraphics[width=1.7in]{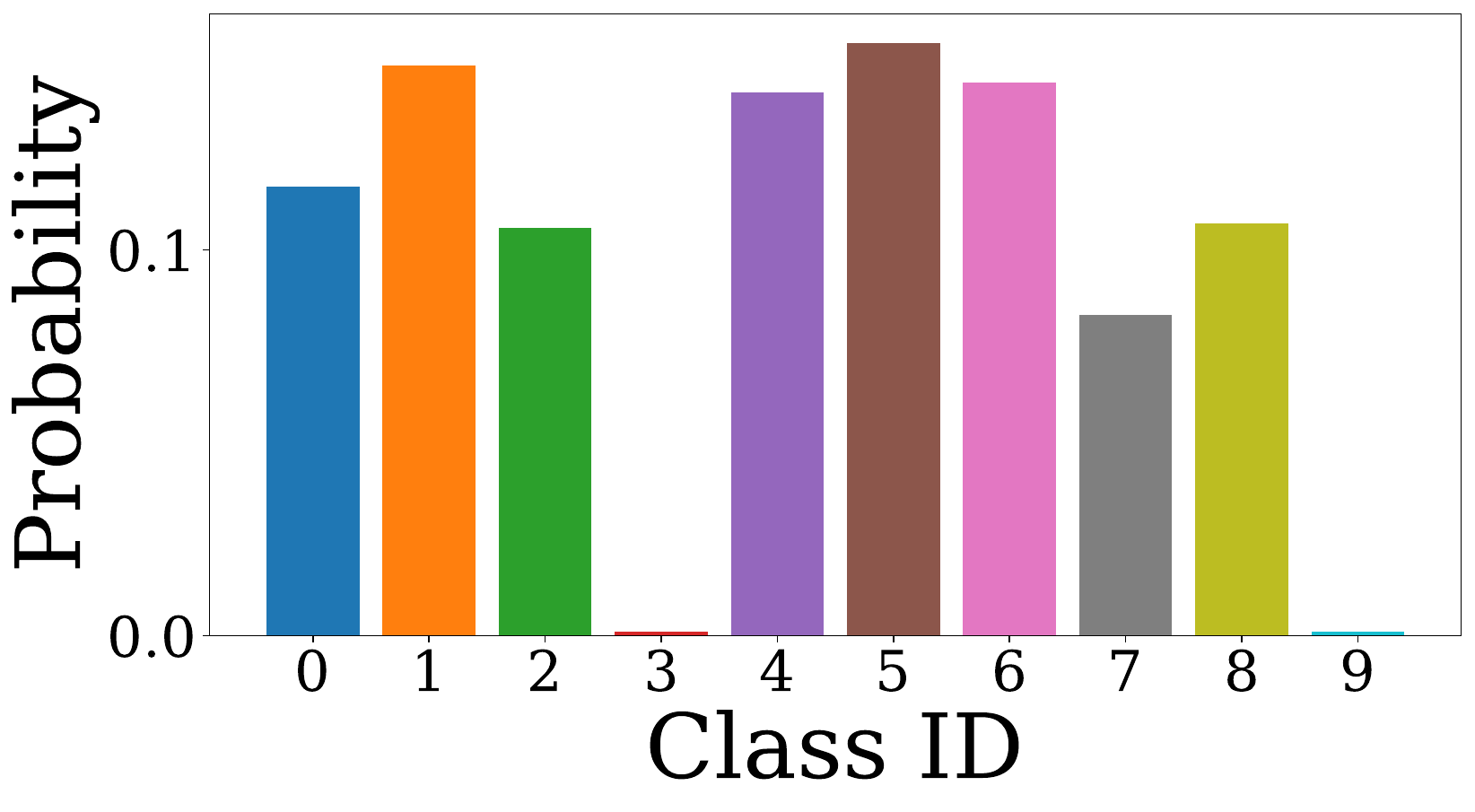}%
\label{fig:Imbalanced_pred_b}}
\caption{Before training (a), the model may not be well-informed and hence shows no strong bias towards any class. After training on an imbalanced dataset (b), the model exhibits a clear bias with significantly varied probabilities for each class. The elevated probability for certain classes suggests that the model has become more confident in these classes likely due to their overrepresentation in the training data. Conversely, the reduced probability for other classes indicates a loss of confidence, which can be interpreted as the model 'forgetting' or failing to recognize these underrepresented classes. }
\label{fig:Imbalanced_pred}
\end{figure}


To mitigate these issues, several strategies have been proposed. Initial attempts focused on enhancing local model training to better align with the global objective. Among these, methods such as parameter sharing \cite{ref1} and characteristics sharing \cite{ref11} of data have been explored to bridge the gap between local and global models. While these approaches aim to enrich local models with broader insights, they inherently increase communication costs and raise privacy concerns, as they necessitate additional data or metadata exchange across the network. Moreover, some researchers have focused on applying generalization theories \cite{ref10}, proposing that enhancing the generalization capability of local models could indirectly address the non-iid challenge. Yet, without direct access to knowledge of the global data distribution, these generalization efforts often fall short, providing limited improvements in overall model performance.

In response to these limitations, the concept of utilizing the global model as a teaching mechanism for local models has gained attention \cite{FedDistill_2022}. This approach leverages the more generalized nature of the global model, aiming to guide local training without the direct exchange of data. However, the effectiveness of such knowledge transfer is markedly reduced in the presence of imbalanced local data distributions. Moreover, initializing local models with the global model's parameters can inadvertently lead to the loss of generalization ability over time, akin to the forgetting phenomenon observed in continuous learning systems. Given this perspective, we ask:\\

\textit{How can we effectively leverage the global model to enhance the learning and generalization ability of local models in a federated learning system, despite the challenges posed by imbalanced local data distribution?
}\\

This study delves into one of the most significant challenges— the phenomenon of model forgetting in the context of non-iid data distribution. Model forgetting, particularly for local models in an FL setup, is often considered as a degradation in the ability to accurately predict underrepresented classes over time. Our investigation reveals that the core of the forgetting issue is primarily rooted in the skewed distribution of local data, where certain classes are represented by fewer samples than others. Such an imbalance leads to insufficient \textquote{positive feedback} during the training process, meaning that the model does not receive enough reinforcement to accurately learn and retain information about these sparsely represented classes.

Figure \ref{fig:Imbalanced_gradient} illustrates the effects of imbalanced class distribution on gradient updates within FL context, using the ResNet-18 architecture trained on CIFAR10 dataset variations. We observe the model's performance across four different training scenarios: \textquote{s1} and \textquote{s3} datasets, which include all classes, and \textquote{s2} and \textquote{s4} datasets, from which class 3 and 9 are removed intentionally to simulate imbalanced class distribution. The analysis of gradients—specifically, how they change in response to the absence or presence of certain classes—reveals that removing classes leads to significant shifts in the model's learning focus. This is depicted through divergent gradient updates for class 3 and 9, indicating the model's \textquote{forgetting} or loss of ability to generalize to those classes that became underrepresented or absent in the training data. 

Following the initial observations, our investigation delves deeper into the mechanisms behind these shifts in learning focus. By analyzing the probability outputs for each class before and after training on both balanced (\textquote{s1}, \textquote{s3}) and imbalanced (\textquote{s2}, \textquote{s4}) datasets as shown in Figure \ref{fig:Imbalanced_pred}, we gain insights into how the model's predictive confidence is affected by the presence or absence of certain classes. This analysis not only confirms the impact of imbalanced datasets on the learning process but also underscores the necessity of addressing this challenge to prevent the degradation of model generalization in FL environments.

We propose a new learning framework, FedDistill, designed to enhance knowledge transfer from the global model to local models in FL, effectively addressing the issue of imbalanced class distribution. 

This approach integrates group distillation with a novel decomposition of the global model into a feature extractor and classifier, targeting the root causes of the 'forgetting' phenomenon in local models. We introduce group distillation, a strategy that segments classes into categories based on their sample abundance in local datasets. This categorization enables a targeted distillation process, where knowledge transfer is customized to bolster the learning of few sample classes, ensuring they receive adequate attention and reinforcement. Furthermore, the global model is decomposed into two components: a feature extractor and a classifier so that local models can learn more effectively from the global knowledge. The feature extractor part of the global model helps local models learn generalized representations of the input data, while the classifier part assists in accurately predicting the class labels. By leveraging the global model as a more effective teacher through group distillation and strategic decomposition, we enable local models to retain their generalization ability over time. This approach mitigates the adverse effects of data imbalance, ensuring that local models do not forget underrepresented classes but instead become more adept at recognizing and classifying them accurately.

Our contributions are as follows:

\begin{itemize}
\item To the best of our knowledge, this is the first in-depth study to identify and analyze the main reason behind the forgetting in local models within the FL framework. We demonstrate that the core issue is the insufficient positive updates for classes with few or no samples during local training, leading to an imbalanced classifier that significantly limits the model's generalization capabilities.


\item We introduce a new learning framework, FedDistill, that incorporates group distillation, tailored to local data distribution, and a decomposed global model approach. This solution is designed to optimize the guidance provided by the global model to local ones, specifically addressing the imbalance in class representation without introducing extra communication overhead or privacy concerns.

\item Through extensive experiments, we comprehensively analyze the efficacy of our proposed components. Our empirical studies on commonly used benchmark datasets reveal that our method achieves state-of-the-art performance, significantly mitigating the forgetting issue and enhancing model generalization across diverse classes.
\end{itemize}

\section{Related Works}
\textbf{Non-IID Federated Learning (FL).} Federated Learning, a paradigm for training models across decentralized datasets without compromising privacy, faces significant challenges in non-iid scenarios. Despite FedAvg's foundational role in FL \cite{ref2}, its performance degrades under non-iid data distributions \cite{ref5}. Subsequent efforts to mitigate these challenges have bifurcated into enhancing global aggregation \cite{ref6,ref7,ref8,ref9} and refining local training methodologies \cite{ref10,ref11,ref12,MOON}. While global-side improvements are constrained by privacy regulations, leading to minimal data-driven modifications, local-side enhancements have flourished. Innovations like FedAlign\cite{ref10} emphasize optimizing local model generalization, while approaches like FedUFO\cite{ref11} and SCAFFOLD\cite{ref12} introduce mechanisms to foster client-agnostic feature learning and update calibration, respectively. Notably, MOON\cite{MOON} leverages contrastive learning to align local model features with the global model, underscoring the evolving landscape of strategies aimed at reconciling local and global learning objectives.

\textbf{Forgetting in Non-iid Federated Learning. }The phenomenon of forgetting, borrowed from lifelong learning, has been recognized in FL, particularly in the context of local training \cite{ref3, ref14, ref16, ref4}. While recent studies have acknowledged its occurrence, the exploration into its causes and the integration of local data distribution into solutions remain nascent. Strategies like reweighting the cross-entropy loss have been proposed to curb forgetting \cite{reweight_softmax}, yet they often do not leverage the global model's potential, inadvertently constraining local models' learning capabilities.

\textbf{Federated Learning with Knowledge Distillation (KD).} In the domain of FL, KD \cite{KD} has emerged as a potent strategy to enhance model performance and generalization by facilitating the transfer of knowledge between models. Originally conceived for model compression, where knowledge from a larger, more complex model is distilled into a smaller, more deployable model, KD has found new applications in FL, particularly in addressing the challenges posed by non-iid data distributions among clients.


KD in the context of FL leverages the architecture where a global model serves as the "teacher" and the local models act as "students." This setup is inherently suitable for FL, where a global model aggregates updates from local models trained on decentralized datasets. The application of KD in FL aims to enhance the ability of local models to learn more generalized features that are representative of the global dataset, thereby improving their performance and the overall efficacy of the FL system.

The process of applying KD in FL can occur in two primary areas: on the local side and on the global side. Local side KD where knowledge from the global model is distilled into local models during their training process. This approach allows local models to benefit from the generalized knowledge of the global model, improving their ability to learn from sparsely represented classes in their local data\cite{FedDistill_2022}. Global side KD where knowledge distillation is applied to enhance the global model's learning from the aggregated updates of local models\cite{ref9,ref28}. Techniques may include generating synthetic data or using unlabeled auxiliary data \cite{FedAux} that mimics the diversity of the global dataset, thereby enabling the global model to learn in a way that reflects the collective knowledge of all participating clients.

While KD offers significant benefits, its application in FL is not without challenges, primarily due to privacy concerns and the need to avoid additional communication overhead. For instance, applying KD on the global side often requires the generation of synthetic data\cite{ref9} or the use of third-party data, which can raise privacy issues or result in data that does not accurately represent the true distribution of client data.

To navigate these challenges, recent innovations have focused on enhancing the local training phase. FedAlign \cite{ref10} suggests that optimizing the generalization ability of local models can yield improvements without necessitating changes to the global aggregation process. Consistency-based methods like FedUFO\cite{ref11} and SCAFFOLD\cite{ref12} aim to encourage local models to learn client-agnostic features, thereby reducing client drift and improving model consistency across different data distributions. Contrastive learning approaches such as MOON\cite{MOON} focus on encouraging clients to learn similar features to the global model, fostering a more uniform representation of knowledge across the network.

Our contribution to this evolving field includes a novel decomposition of knowledge distillation into three parts: true class, few-sample class, and rich-sample class, directly addressing the challenge of imbalanced data distribution. Furthermore, by decomposing both the global and local models, we apply KD in a manner that effectively prevents drift from the global model to local models, ensuring a more balanced and representative learning process. Notably, our method emphasizes improvements on the local side, without any additional communication overhead and privacy concerns typically associated with global-side enhancements.

\section{Method}

\subsection{Preliminary} \label{sec:pre}

In the FL setting, following the FedAvg algorithm, we consider a set of clients $S$, where each client $s \in S$ possesses a private dataset $\mathcal{D}_s$. Each client has an associated image classification model represented by parameters $\theta_s$. The global model, shared among all clients, is denoted by $\theta_g$, and the collective dataset from all clients is represented as $\mathcal{D} = \bigcup_{s \in S} \mathcal{D}_s$. The FL process is structured into $T$ communication rounds. In each round $t$, a subset of clients $S^{(t)}$ is selected, and the global model from the previous round $\theta_{g}^{(t-1)}$ is distributed to each selected client $s \in S^{(t)}$. This model initializes their local model for the current round as $\theta_{s}^{(t)} = \theta_{g}^{(t-1)}$, upon which local training is conducted.

The aggregation of trained local models to update the global model is performed by averaging their weights, given by:
\begin{equation}
\theta_{g}^{(t)} = \frac{\sum_{s \in S^{(t)}} |\mathcal{D}_s| \theta_{s}^{(t)}}{\sum_{s' \in S^{(t)}} |\mathcal{D}_{s'}|}
\end{equation}

The overall objective is to minimize the empirical loss of the global model across the entire dataset, formalized as:
\begin{equation}
\min_{\theta_g} \mathbb{E}_{(\boldsymbol{x},y) \sim \mathcal{D}} \left[ \mathcal{L} \left( f(\boldsymbol{x}; \theta_g), y \right) \right]
\end{equation}
where $\mathcal{L}$ denotes the loss function, $(\boldsymbol{x}, y)$ are the input image and corresponding ground-truth label pairs, and $f(\boldsymbol{x}; \theta_g)$ represents the mapping from the input space $\mathcal{X}$ to the label space $\mathcal{Y}$, parameterized by the global model $\theta_g$.






\subsection{Understanding Forgetting in Federated Learning} \label{sec:forget}

The phenomenon of forgetting within local models during Federated Learning (FL) training sessions significantly hampers their ability to generalize, particularly for classes that are underrepresented in their datasets. This subsection delves into the mechanisms contributing to this issue and proposes a novel perspective on mitigating its impact through the strategic use of the global model.

\begin{figure}[!t]
\centering
\includegraphics[width=3.5in]{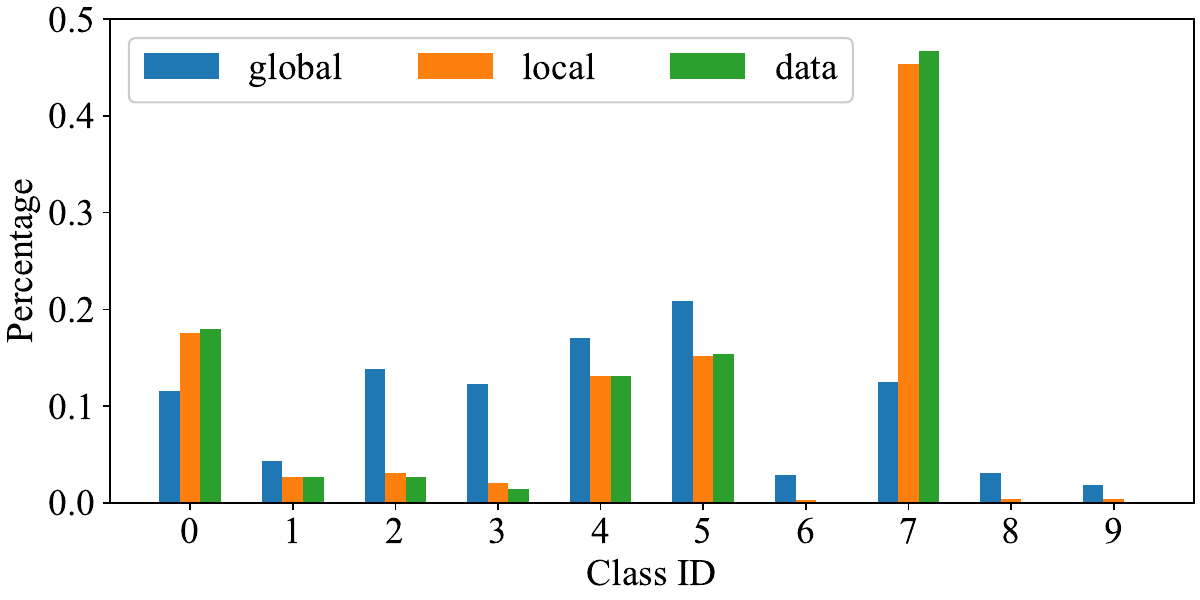}
\caption{Illustration of softmax output distribution disparities between global and local models under imbalanced class distribution on client 0, highlighting the local model's inclination towards fitting its specific dataset and the global model's balanced approach.}
\label{fig:Imbalanced_client_0}
\end{figure}

Previous investigations have identified client drift, backward transfer (BwT), and unbalanced loss as indicators of forgetting during local training sessions \cite{ref1, ref3, ref4}. However, these studies fall short of pinpointing the precise mechanism of forgetting induced by non-iid data distributions. Our analysis reveals that the core issue lies within the cross-entropy loss function, a common choice for classification tasks in FL.

\textbf{The role of cross-entropy loss.} For a given classification model $\theta$, which comprises a feature extractor $E$ ($E(\boldsymbol{x}): \mathcal{X} \rightarrow \mathbb{R}^d$, mapping inputs to a feature space) and a classifier $FC$ ($FC(E(\boldsymbol{x}))= W^T E(\boldsymbol{x})$, mapping features to class probabilities), we can express the output probability for a class $c \in \mathcal{C}$ as:

$$q_{c}(\boldsymbol{x})=\frac{\exp \left(\boldsymbol{w}_{c}^{T} E(\boldsymbol{x})\right)}{\sum_{k}^{|\mathcal{C}|} \exp \left(\boldsymbol{w}_{k}^{T} E(\boldsymbol{x})\right)},$$

with the cross-entropy loss calculated as:

$$\sum_{c \in \mathcal{C}} p_{c} \log \left(q_{c}(\boldsymbol{x})\right),$$

where $p_c$ represents the ground truth. The derivative of the loss with respect to the weight of the ground-truth class $c$ is positive, promoting learning for this class:

$$\frac{\partial \log \left(q_{c}\right)}{\partial \boldsymbol{w}_{c}}=\left(1-q_{c}\right) E(\boldsymbol{x}),$$

while for non-true classes $\bar{c}$, it's negative, potentially decreasing their weight:

$$\frac{\partial \log \left(q_{c}\right)}{\partial \boldsymbol{w}_{\bar{c}}}=-q_{\bar{c}} E(\boldsymbol{x}).$$

This differential treatment can lead to the forgetting of underrepresented classes, as their weights diminish over time due to insufficient positive feedback. 

\textbf{Fitting local data and losing generalization.} Fig. \ref{fig:Imbalanced_client_0} showcases a FedAvg scenario where the local model's softmax output distribution significantly deviates towards classes overrepresented in its dataset, compromising its ability to generalize. In contrast, the global model maintains a more balanced output distribution despite the local data imbalance.

\textbf{Proposed mitigation through global model guidance.} In response to these challenges, our approach advocates for a re-envisioned application of the global model in guiding local training. Unlike prior efforts that solely focus on optimizing local models' generalization abilities or sharing parameters/data characteristics, we propose a method that incorporates class imbalanced data distribution directly into the training process. This involves using the global model to provide targeted guidance to local models, enhancing their ability to retain knowledge of underrepresented classes without introducing additional communication burdens or privacy concerns.

By situating the global model as a more generalized teacher and decomposing it into distinct feature extractor and classifier components, we offer a nuanced strategy that not only addresses the forgetting issue but also respects the federated learning paradigm's foundational principles of efficiency and privacy.

\subsection{FedDistill Framework}
The FedDistill framework introduces an advanced approach to Federated Learning (FL) by addressing the challenges posed by non-iid data distributions across clients. At the heart of FedDistill is the innovative use of group distillation (GD) Loss, which modifies the traditional knowledge distillation process to better suit the federated context. This section outlines the FedDistill framework, emphasizing the strategic components designed to enhance local model performance through global model insights.


\begin{figure}[!t]
\centering
\includegraphics[width=3.5in]{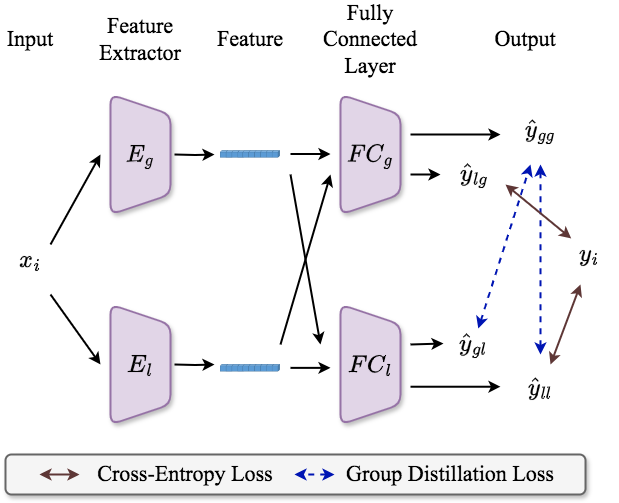}
\caption{Overview of the FedDistill framework, detailing the interplay between global and local models' feature extractors and classifiers for an input $x_i$. Specifically, $x_i$ is the input for the global feature extractor $E_g$ and local feature extractor $E_l$, respectively. After that, the features will be input into the classifiers $FC_g$ and $FC_l$ of both the global and the local model. $\hat{y}_{gg}$, $\hat{y}_{gl}$ denotes the output of global and local classifier with the global feature, respectively, and $\hat{y}_{lg}$, $\hat{y}_{ll}$ denotes the output of global and local classifier with the local feature, respectively. $y_i$ is the real output.}
\label{fig:framework}
\end{figure}


Fig. \ref{fig:framework} illustrates the framework's structure, where both global and local models contribute to a cross-learning environment. This setup not only leverages the generalized capabilities of the global model but also specifically targets the improvement of local models' generalization and robustness against class imbalance.

\textbf{Group distillation loss.} In FL, where data is inherently non-iid across clients, traditional KD methods, which rely on KL divergence to measure and minimize the discrepancy between the teacher (global) and student (local) models, often fall short. This shortfall is particularly evident in handling the class imbalance problem within local datasets.

\begin{figure}[!t]
\centering
\includegraphics[width=3.5in]{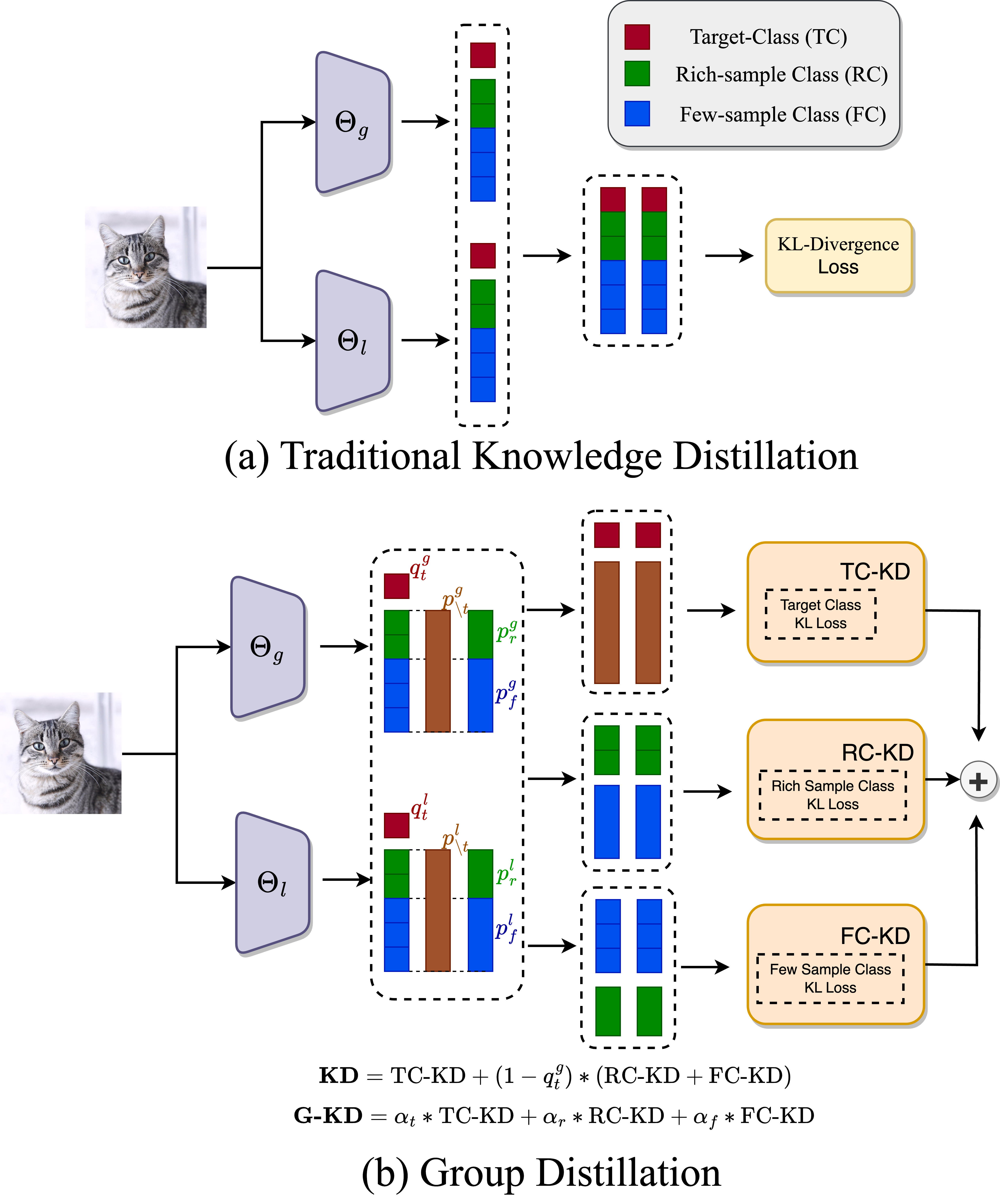}
\caption{Illustration of the traditional KD and our G-KD. We reformulate the traditional KD into three parts: (1) The target class KL loss (TC-KD), which has been discussed in DKD\cite{ref20}. (2) The rich sample KL loss (RC-KD), denotes the KL loss for the rich sample classes that were highly represented by the local modal. (3) The few sample KL loss (FC-KD), denotes the KL loss for the few sample classes that were underrepresented by the local model. By separating the classes, we intended to accommodate the imbalance in the local dataset distribution by adjusting the weight ($\alpha_t, \alpha_r, \alpha_f$) correspondingly.}
\label{fig:GKL}
\end{figure}

Our novel GD Loss approach modifies the traditional use of KL divergence to better accommodate the unique distributional characteristics of FL. By segmenting the local data into three categories—true-class, rich-sample class, and few-sample class—based on sample abundance. Specifically, we introduce a parameter termed the few-sample threshold, denoted as $\gamma \in [0,1]$. Each client's data is then classified into rich-sample classes if the sample proportion exceeds $\gamma$; otherwise, it is categorized into few-sample classes. we apply a differentiated distillation technique that recalibrates the focus of local models towards achieving a more balanced learning outcome across all classes. A visual demonstration of GD Loss is shown in Figure \ref{fig:GKL}.

To detail, we reformulate the KL divergence to encapsulate the differentiated treatment of each class group, thus ensuring a more equitable learning process across classes with varying levels of representation:

\begin{equation}
\begin{aligned}
\operatorname{GD}\left(\boldsymbol{q}^{g} \| \boldsymbol{q}^{l}\right) = &\ \alpha_t \operatorname{TC-KD} (\boldsymbol{q}^{g}, \boldsymbol{q}^{l}) \\
&+ \alpha_r \operatorname{RC-KD} (\boldsymbol{q}^{g}, \boldsymbol{q}^{l}) \\
&+ \alpha_f \operatorname{FC-KD} (\boldsymbol{q}^{g}, \boldsymbol{q}^{l}),
\end{aligned}
\end{equation}


where $\operatorname{TC-KD}$, $\operatorname{RC-KD}$, and $\operatorname{FC-KD}$ represent the components of our GD Loss corresponding to true-class, rich-sample class, and few-sample class, respectively. For classes have    Each component is scaled by its respective weighting factor ($\alpha_t$, $\alpha_r$, and $\alpha_f$) to ensure that the distillation process is finely tuned to the specific learning needs and challenges posed by the class distribution within each local dataset. More specifically, their formulations are as follows:

\begin{equation}
\operatorname{TC-KD}\left(\boldsymbol{q}^{g} \| \boldsymbol{q}^{l}\right) =  
q^{g}_{t}\log \left( \frac{q^g_t}{q^l_t} \right) + p^{g}_{\backslash t}\log \left( \frac{p^g_{\backslash t}}{p^l_{\backslash t}} \right)
\end{equation}

where $q_t$ denotes the logit of the true class, and $p_{\backslash t}$ denotes the logit sum of the not-true classes.

\begin{equation}
\operatorname{RC-KD}\left(\boldsymbol{q}^{g} \| \boldsymbol{q}^{l}\right) =  
\left( \sum_{i\in \mathcal{C}_{r} \backslash \{t\}} \tilde{q}^g_i \log \left( \frac{\tilde{q}^g_i}{\tilde{q}^l_i} \right) + \tilde{p}^g_f \log \left( \frac{\tilde{p}^g_f}{\tilde{p}^l_f} \right) \right)
\end{equation}

where $\tilde{q}_i = \frac{q_i}{p_{\backslash t}}$, $\tilde{p}_f = \frac{\sum_{i\in \mathcal{C}_{f} \backslash \{t\} } q_i }{p_{\backslash t}}$, and $\mathcal{C}_{r}$, $\mathcal{C}_{f}$ denotes the rich-sample and few-sample classes respectively.

\begin{equation}
\operatorname{FC-KD}\left(\boldsymbol{q}^{g} \| \boldsymbol{q}^{l}\right) =  
\left( \sum_{i\in \mathcal{C}_{f} \backslash \{t\}} \tilde{q}^g_i \log \left( \frac{\tilde{q}^g_i}{\tilde{q}^l_i} \right) + \tilde{p}^g_r \log \left( \frac{\tilde{p}^g_r}{\tilde{p}^l_r} \right) \right)
\end{equation}

where $\tilde{p}_r = \frac{\sum_{i\in \mathcal{C}_{r} \backslash \{t\} } q_i }{p_{\backslash t}}$. When $\alpha_t$, $\alpha_r$, and $\alpha_f$ are set to 1, $p_{\backslash t}$, $p_{\backslash t}$ respectively, GD Loss is equivalent to KL divergence Loss.

Our reconceptualization of KL divergence, pivotal to the GD Loss, addresses the prevalent class imbalance issue in FL, ensuring effective knowledge transfer from the global to local models.  By ensuring that the distillation process respects and responds to the unique class distribution of each local dataset, we facilitate a more balanced and effective knowledge transfer from global to local models. This strategy not only counteracts the adverse effects of non-iid data on model performance but also significantly enhances the robustness and generalization capabilities of FL systems.

\textbf{Feature Extraction and Classifier Enhancement.} 
 Since the global model can be viewed as a more generalized model compared to local ones \cite{MOON}, we assume that the feature extractor and the classifier of the global model are also generalized. To preserve the generalization ability, we intend to let global teach the local model, by utilizing the group distillation.

\begin{equation}
    \mathcal{L}_{\text{L}} = \operatorname{GD}(\hat{y}_{gg} \| \hat{y}_{ll})
\end{equation}

where $\hat{y}_{gg}, \hat{y}_{ll}$ are the output of the global and local model with the same input $\boldsymbol{x}$. The GD process aims to align the local model's predictions more closely with those of the global model, thereby leveraging the global model's generalization capabilities to mitigate biases and imbalances within the local model's learning process.

Acknowledging the influence of imbalanced local data on the classifier's performance, an innovative step is taken to integrate the global classifier in evaluating the local feature extractor. By processing local features through the global classifier ($FC_g$), we obtain a less biased prediction ($\hat{y}_{lg}$), which serves as a more reliable indicator of the local feature extractor's ($E_l$) generalization performance. This process aims to ensure that the local feature extractor remains discriminative and effective across both biased and unbiased classifiers, with the goal articulated through the following loss function.

\begin{equation}
    \mathcal{L}_{\text{E}} = \operatorname{CE}(\hat{y}_{lg}, y)
\end{equation}

focusing on minimizing the cross-entropy (CE) loss between $\hat{y}_{lg}$ and the ground-truth label $y$.

Further, to address biases within the local classifier, we explore the concept of preserving the ranking information conveyed by the global classifier. This is achieved by processing features extracted by the global feature extractor ($E_g$) through the local classifier ($FC_l$), generating a prediction ($\hat{y}_{gl}$) that reflects the performance of an unbiased feature extractor when coupled with a biased classifier. This strategy, aimed at preserving ranking information, is quantified through the minimization of Group Distillation between $\hat{y}_{gl}$ and the global model's predictions ($\hat{y}_{gg}$).

\begin{equation}
    \mathcal{L}_{\text{FC}} = \operatorname{GD}(\hat{y}_{gg} \| \hat{y}_{gl})
\end{equation}

In conjunction with these targeted interventions, the original CE loss between the local model's predictions and the ground-truth labels is maintained, ensuring the local model's foundational predictive capabilities are not compromised. The composite loss function, incorporating CE loss and the components designed to enhance feature extraction and classification, is presented as follows:

\begin{equation}
    \mathcal{L} = \operatorname{CE}(\hat{y}_{ll}, y) + \beta_{\text{L}} \mathcal{L}_{\text{L}} + \beta_{\text{E}} \mathcal{L}_{\text{E}} + \beta_{\text{FC}} \mathcal{L}_{\text{FC}}
\end{equation}

with $\beta_{\text{L}}$, $\beta_{\text{E}}$, and $\beta_{\text{FC}}$ serving as weighting factors for the respective loss components. This structured approach ensures a comprehensive and balanced enhancement of local models' capabilities, addressing the nuances of non-iid data distribution and class imbalance within the federated learning paradigm.

\section{Experiments and Results}
\subsection{Experimental Setup}
\begin{figure*}[!t]
\centering
\subfloat[CIFAR10: $\alpha=0.5$]{\includegraphics[width=1.5in]{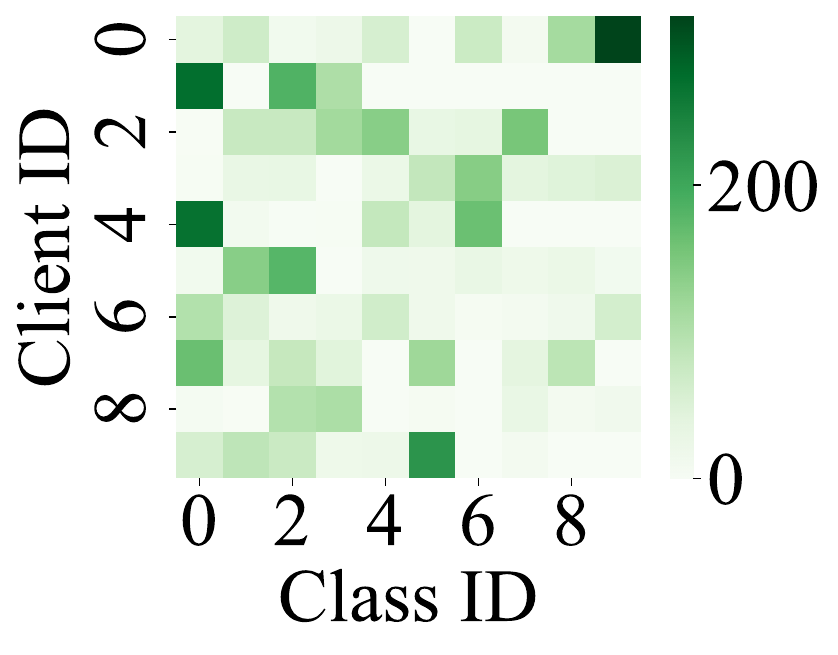}%
\label{fig:cifar10_a0.5}}
\hfil
\subfloat[CIFAR10: $\alpha=0.3$]{\includegraphics[width=1.5in]{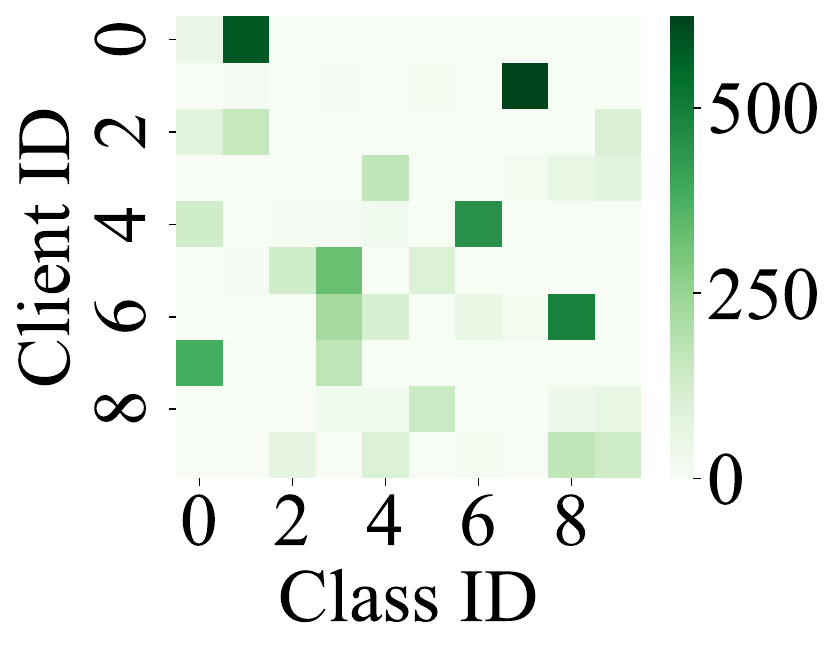}%
\label{fig:cifar10_a0.3}}
\hfil
\subfloat[CIFAR10: $\alpha=0.1$]{\includegraphics[width=1.5in]{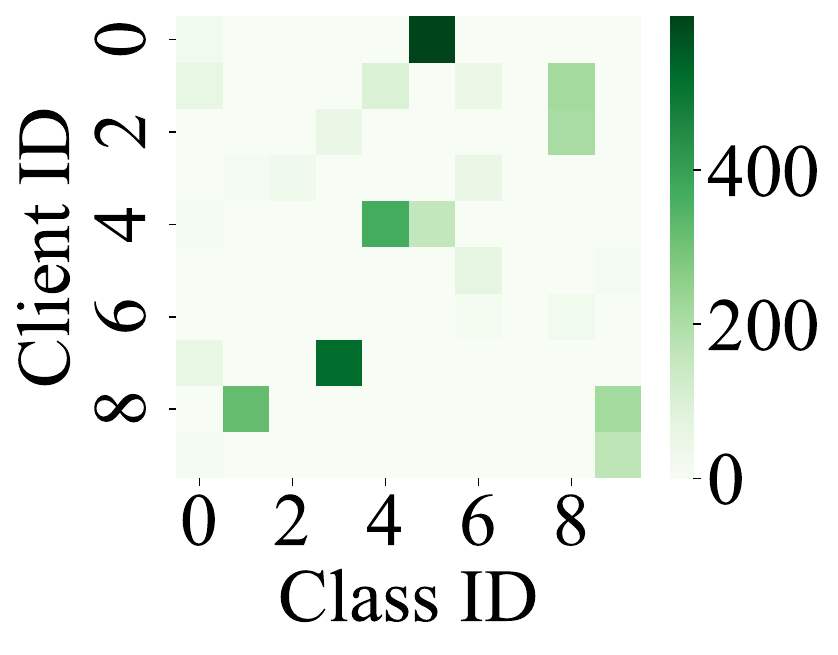}%
\label{fig:cifar10_a0.1}}
\hfil
\subfloat[MNIST: $\alpha=0.1$]{\includegraphics[width=1.5in]{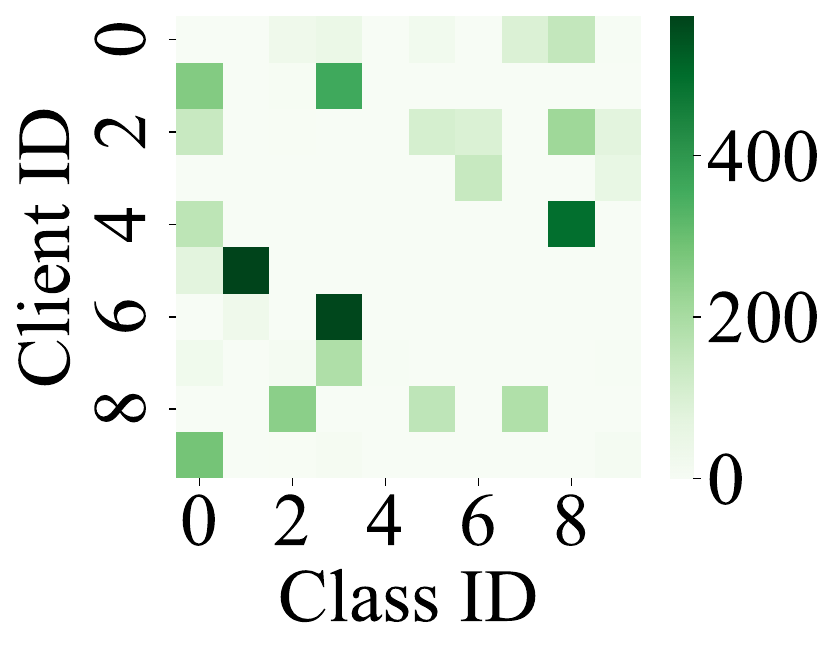}%
\label{fig:mnist_a0.1}}
\hfil
\subfloat[CIFAR100: $\alpha=0.1$]{\includegraphics[width=7in]{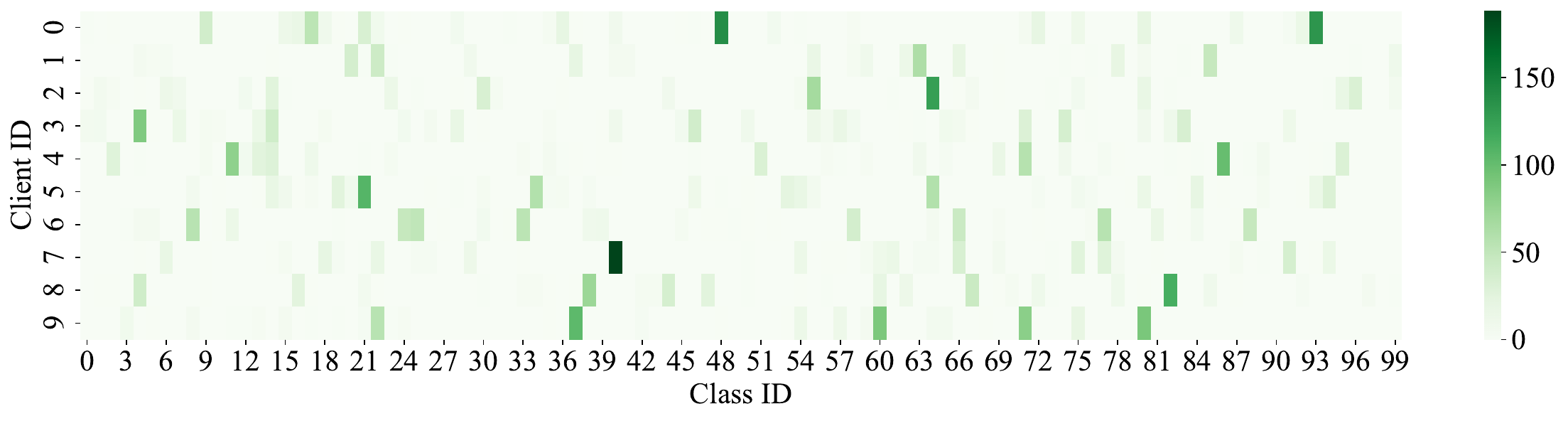}%
\label{fig:cifar100_a0.1}}

\caption{Data distribution with different non-iid levels in the first ten clients for all datasets. (a) - (c) demonstrate how non-iid level affects the data sparsity.}
\label{fig:data_dist}
\end{figure*}

\textbf{Datasets.} We conduct experiments on commonly used datasets in FL, which are MNIST, CIFAR10, and CIFAR100. We also adopt a commonly used heterogeneous dataset partition method \cite{ref22, ref9, MOON} using Dirichlet distribution $\operatorname{Dir}(\alpha)$ to assign the label distribution among to simulate the non-iid data distribution across clients. The $\alpha$ is $0.1$ for MNIST and CIFAR100, and $\{0.1,0.3,0.5\}$ for CIFAR10 to show the performance of our method under different non-iid levels. As shown in Fig. \ref{fig:data_dist}, we visualize the data distribution with responses to classes in the first ten clients. From Fig. \ref{fig:cifar10_a0.5} to Fig. \ref{fig:cifar10_a0.1}, A clear increase of data sparsity is observed. For all datasets, we set 100 clients and sampled 10\% of them for each communication round. There are a total of 100 communication rounds for each dataset. More details of those datasets are shown in Table. \ref{tab:datasets}. 

\begin{table}[!t]
\caption{Configurations of datasets}\label{tab:datasets}
\centering
\begin{tabular}{cccc}
\hline
\textbf{Datasets} & \textbf{CIFAR10} & \textbf{CIFAR100} & \textbf{MNIST}   \\
\hline
Dataset Classes & 10 & 100 & 10  \\
Training Instance & 50,000 & 50,000 & 60,000  \\
Test Instance & 10,000 & 10,000 & 10,000,  \\
Model & SimpleCNN & Resnet18 & SimpleCNN  \\
Local epochs & 10 & 10 & 10  \\
Batch Size & 64 & 64 & 64  \\
Number of Clients & 100 & 100 & 100  \\
Sample ratio & 10\% & 10\% & 10\%  \\
Non-iid Level ($\alpha$) & $\{0.1, 0.3, 0.5\}$ & 0.1 & 0.1  \\
\hline
\end{tabular}
\end{table}


\textbf{Training and Hyperparameters Configurations.} Following existing works \cite{MOON, ref3, ref9}, we adopt SimpleCNN and resnet18 as the classification model. We apply resnet18 in CIFAR100 to demonstrate our method is applicable in modern CNN-based architecture. For all datasets, the number of local epochs is 10 and the batch size is 64. We take 0.01 as the learning rate for SimpleCNN and we adopt SGD as optimizer with momentum as 0.9 and weight-decay as $1e^{-5}$. We first search an optimal $\alpha_t, \alpha_r$ and $\alpha_f$ from $\{0.0, 0.1, 0.2, \ldots, 1.0\}$. We will freeze the $\alpha_t, \alpha_r$ and $\alpha_f$ and search the $\beta_{\text{L}}$ from $\{0.5, 0.8, 1.0, 1.1, 1.2,1.4\}$, and $ \beta_{\text{E}}$ $\beta_{\text{FC}}$ from $\{0.1, 0.2, 0.3, 0.4, 0.5\}$, respectively. For the few-sample threshold $\gamma$, we empirically set it to $\frac{1}{\vert \mathcal{C} \vert}$ for each dataset.

\textbf{Baselines.} In our work, the adoption of widely recognized baselines such as FedAvg, FedProx, SCAFFOLD, MOON, and the more recent FedNTD, underscores a comprehensive approach to evaluating the effectiveness of our proposed solution in the context of FL under non-IID data distributions. Each of these baselines represents a significant contribution to the FL landscape, addressing various aspects of the challenges posed by non-IID data:

\begin{itemize}
\item \textbf{FedAvg} \cite{ref2} is the foundational algorithm in FL that aggregates model updates from a diverse set of clients. It serves as the standard against which improvements in handling non-IID data are measured.
\item \textbf{FedProx} \cite{ref23} introduces modifications to FedAvg to better handle statistical heterogeneity and system challenges, making it a relevant comparison for assessing enhancements in model robustness and stability.
\item \textbf{SCAFFOLD} \cite{ref12} tackles the variance in updates due to non-IID data by correcting the client drift, providing a direct comparison for evaluating improvements in model convergence and accuracy.
\item \textbf{MOON} \cite{MOON} leverages contrastive learning to mitigate the effects of data heterogeneity, offering a novel perspective on enhancing model generalization across clients with diverse data distributions.
\item \textbf{FedNTD} \cite{ref3} specifically addresses the issue of forgetting in FL by incorporating concepts from lifelong learning, making it an apt benchmark for comparing the efficacy of our method in maintaining model performance over time.
\end{itemize}

For the implementation of baseline methods, we followed the original implementation. More specifically, we set $\mu=0.1$ for FedProx, $\mu=1.0$, $\tau=1.0$ for MOON and $\mu=1.0$, $\tau=1.0$ for FedNTD where $\mu$ denotes the coefficient of the losses proposed by these methods, and $\tau$ denotes the temperature of the distillation. There is no hyperparameter to tune for FedAvg and SCAFFOLD. For all the baseline methods, we adopted the same network, and the same training setting(learning rate, optimizer settings, batch size, and local epochs) as our method.

By comparing our approach against these baselines, we can effectively demonstrate its advantages in terms of de-biasing local models, enhancing generalization, and reducing the impact of class imbalance in FL environments.

\textbf{Evaluation.} For all the results, we take the top-1 accuracy as the primary metric. We take the average of three separate runs with random seeds in 2022, 2023, and 2024 for the convenience of reproduction. Apart from the top-1 accuracy, we also introduced Backward Transfer $\mathcal{F}$ proposed in \cite{fm} as a measurement of forgetting of models. The formulation of $\mathcal{F}$ is as follows:

$$\mathcal{F} = \frac{1}{\vert \mathcal{C} \vert} \sum_{c \in \mathcal{C}} \underset{t \in {\{1,\ldots, T-1\}}}{\operatorname{max}} \mathcal{A}^{t}_{c} - \mathcal{A}^{T}_{c}$$

where $\mathcal{A}^{t}_{c}$ denotes the accuracy of the global model on class $c$ in the $t$-th communication round, and $\mathcal{A}^{T}_{c}$ is the same of the last round. All methods adopt the same model architecture and training setup specified in Table. \ref{tab:datasets}.


\subsection{Performance Analysis}
\begin{table*}[!t]
\caption{Top-1 Accuracy (\%) of all the baselines and our method on all datasets. The number inside the bracket is the Forgetting measure $\mathcal{F}$.} \label{tab:top-1 acc}
\centering
\scalebox{1.3}{
\begin{tabular}{cccccc}
\hline
\multirow{2}*{Methods} & \multirow{2}*{MNIST} & \multicolumn{3}{c}{CIFAR10} & \multirow{2}*{CIFAR100} \\
~ & ~ & $\alpha=0.1$ & $\alpha=0.3$ & $\alpha=0.5$ & ~  \\
\hline
\hline
FedAvg & 77.97{\tiny(0.2191)} & 35.37{\tiny(0.6357)} & 49.68{\tiny(0.4318)} & 55.6{\tiny(0.3308)} & 32.19{\tiny(0.3463)} \\
\hline 
FedProx & 65.81{\tiny(0.3413)} & 34.29{\tiny(0.6300)} & 41.77{\tiny(0.4743)} & 44.4{\tiny(0.4135)} & 29.48{\tiny(0.2)} \\
SCAFFOLD & 79.98{\tiny(0.1994)} & 36.03{\tiny(0.6295)} & 50.17{\tiny(0.4261)}  & 55.33{\tiny(0.3324)} & 32.35{\tiny(0.3507)}\\
MOON & 79.45{\tiny(0.205)} & 34.83{\tiny(0.6421)} & 49.48{\tiny(0.4278)} & 55.72{\tiny(0.3270)} & 32.45{\tiny(0.3563)} \\
FedNTD & 89.46{\tiny(0.0976)} & 47.11{\tiny(0.4624)} & 54.85{\tiny(0.3026)} & 57.04{\tiny(0.2456)} & 35.33{\tiny(0.1533)} \\
\hline
FedDistill (\textbf{Ours}) & \textbf{90.35}{\tiny(0.0903)} & \textbf{47.78}{\tiny(0.4611)} & \textbf{56.72}{\tiny(0.2656)} & \textbf{58.88}{\tiny(0.2008)} & \textbf{36.57}{\tiny(0.1493)} \\
\end{tabular}}
\end{table*}

\begin{table*}[!t]
\caption{Communication Efficiency. It demonstrates the number of communication rounds that each approach achieves the final accuracy of FedAvg on the same dataset.}\label{tab:Com-Eff}
\centering
\scalebox{1.3}{
\begin{tabular}{cccccc}
\hline
\multirow{2}*{Methods} & \multirow{2}*{MNIST} & \multicolumn{3}{c}{CIFAR10} & \multirow{2}*{CIFAR100} \\
~ & ~ & $\alpha=0.1$ & $\alpha=0.3$ & $\alpha=0.5$ & ~   \\
\hline
\hline
FedAvg & 100 & 100 & 100 & 100 & 100 \\
\hline 
FedProx & N/A & 91  & N/A & N/A & N/A \\
SCAFFOLD & 41 & 94 & 59 & 62 & 90 \\
MOON & 41 & N/A & 59 & 73 & 87 \\
FedNTD & 19 & 34 & 33 & 48 & 60 \\
\hline
FedDistill (\textbf{Ours}) & 16 & 18 & 25 & 43 & 49\\
\end{tabular}}
\end{table*}

Our comprehensive evaluation, as detailed in Table \ref{tab:top-1 acc}, showcases the top-1 accuracy and forgetting measure ($\mathcal{F}$) across various datasets for our FedDistill approach against established baselines such as FedAvg, FedProx, SCAFFOLD, MOON, and FedNTD. Notably, our method demonstrates superior performance in terms of both accuracy and the mitigation of forgetting, indicative of its robustness and efficacy in non-IID Federated Learning environments.

\textbf{Comparative Performance.}
The results show FedDistill's ability to significantly enhance model accuracy across all evaluated datasets, including MNIST, CIFAR10 with varying degrees of data heterogeneity ($\alpha=0.1, 0.3, 0.5$), and CIFAR100. Specifically, our method achieves the highest top-1 accuracy of 90.35\% on MNIST and notable improvements in CIFAR10 and CIFAR100, surpassing the next best method, FedNTD, by a margin that highlights the effectiveness of our distillation strategy.

The forgetting measure ($\mathcal{F}$) further validates our method's capability to retain learned knowledge more effectively than competing approaches. FedDistill exhibits the lowest $\mathcal{F}$ values across all datasets, affirming its superiority in addressing the critical challenge of model forgetting in FL. This is particularly evident in the CIFAR10 dataset, where our approach not only enhances accuracy but also significantly reduces the extent of forgetting compared to other methods.

Our experimental findings also reveal that FedDistill offers faster convergence rates, underscoring the efficiency of our model distillation process. This rapid convergence is coupled with stability across different network architectures, showcasing our method's versatility. Specifically, the experiments on CIFAR100 and the comparative analysis between network architectures (resnet18 vs. simpleCNN) demonstrate that our method is adaptable to varying class numbers and network complexities.

\begin{figure*}[!t]
\centering
\subfloat[FedAvg]{\includegraphics[width=.3\textwidth]{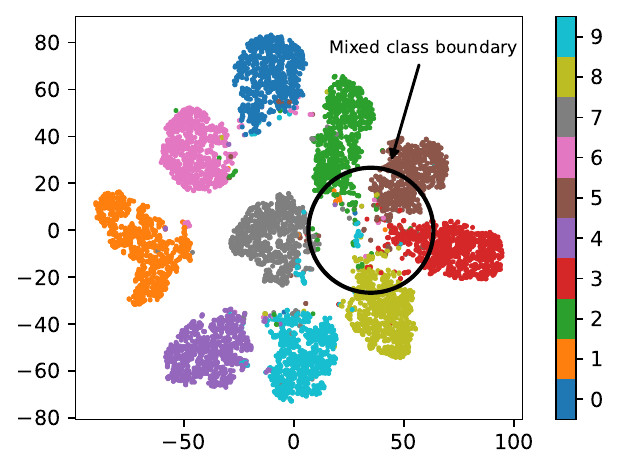}%
\label{fig:tsne-fedavg}}
\hfil
\subfloat[FedProx]{\includegraphics[width=.3\textwidth]{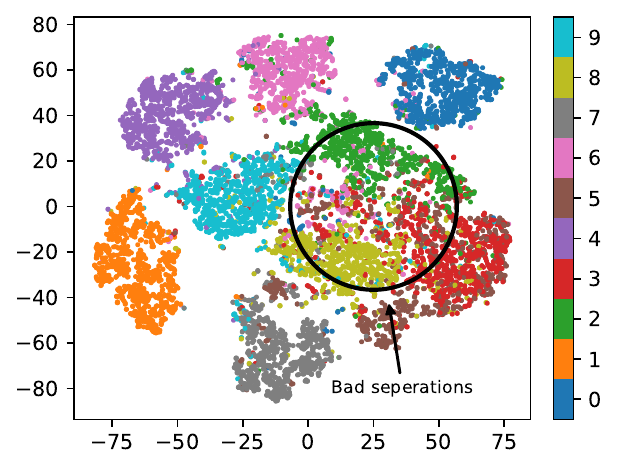}%
\label{fig:tsne-fedprox}}
\hfil
\subfloat[SCAFFOLD]{\includegraphics[width=.3\textwidth]{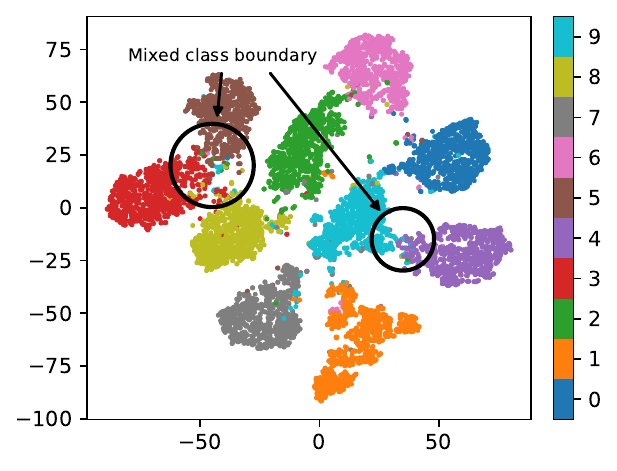}%
\label{fig:tsne-scaffold}}
\hfil
\subfloat[MOON]{\includegraphics[width=.3\textwidth]{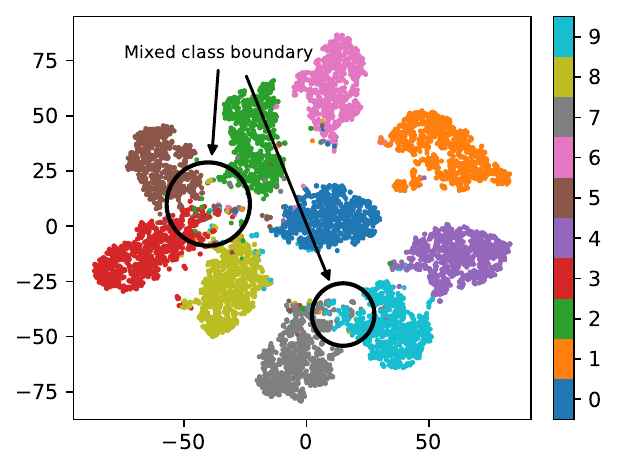}%
\label{fig:tsne-moon}}
\hfil
\subfloat[FedNTD]{\includegraphics[width=.3\textwidth]{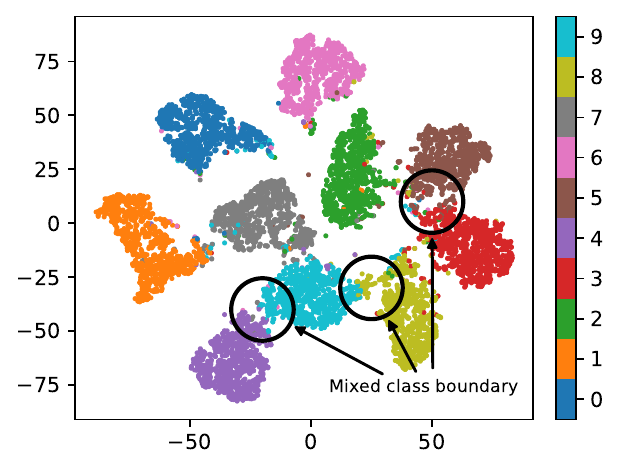}%
\label{fig:tsne-fedntd}}
\hfil
\subfloat[FedDistill(Ours)]{\includegraphics[width=.3\textwidth]{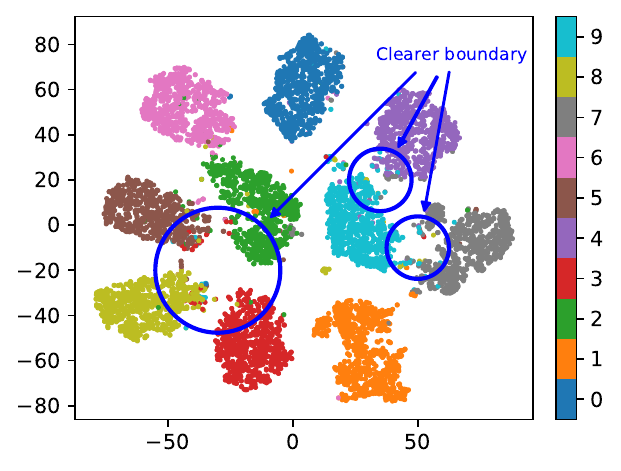}%
\label{fig:tsne-fedcdl}}

\caption{The t-SNE visualization of the features extracted by the feature-extractor of different models from MNIST with non-iid level as 0.1. The black circles denote mixed class boundaries and the blue circles denote clearer class boundaries. Compared to baseline methods, our model yield a clearer class boundary, demonstrating that our model learn a more discriminative feature extractor.}
\label{fig:t-sne}
\end{figure*}

Fig. \ref{fig:t-sne} showcases the t-SNE visualization of features extracted by the feature-extractor of different models from MNIST with non-iid level as 0.1. The mixed class boundaries were highlighted with black circles on the graph, while the blue circle denotes a clearer class boundary compared to other methods. From the graph, we found that: (1) Almost all baselines are struggling to classify class 2 (green), 3 (red), 5 (brown), and 8 (yellow) (2) There are also mixed class boundaries between class 7 (grey) and 9 (blue), 4 (purple) and 9, and 8 and 9, which is MOON, SCAFFOLD and FedNTD, and FedNTD, respectively. (3) Our method has clear boundaries among those classes. It shows that features extracted by baseline models often failed to have clear class boundaries even with a high top-1 accuracy (89.46), while our method could extract more discriminative features, indicating that our method is capable of learning a better feature-extractor.

The key observations are as follows. (1) Our method effectively de-biases local models by leveraging global model insights, significantly enhancing local model generalization and reducing the impact of non-IID data distributions. (2) The direct correlation between the forgetting measure ($\mathcal{F}$) and performance underscores the importance of addressing forgetting in improving FL algorithm performance. FedDistill's lower $\mathcal{F}$ values across datasets highlight its effectiveness in mitigating forgetting. (3) Our method is able to extract more discriminative features from the data, which is aligned with our motivation of feature extraction enhancement. (4) The consistent performance of our method across diverse datasets and network architectures attests to its robustness and adaptability, making it a promising solution for a wide range of FL applications.

In summary, the empirical evidence from our analysis firmly establishes the FedDistill method as a significant advancement in FL, particularly in tackling the challenges posed by non-IID data distributions. By effectively de-biasing local models and mitigating forgetting, our approach not only enhances model performance but also contributes to the development of more equitable and efficient distributed learning systems.





\subsection{Communication Efficiency}

\begin{figure*}[!t]
\centering
\includegraphics[width=7in]{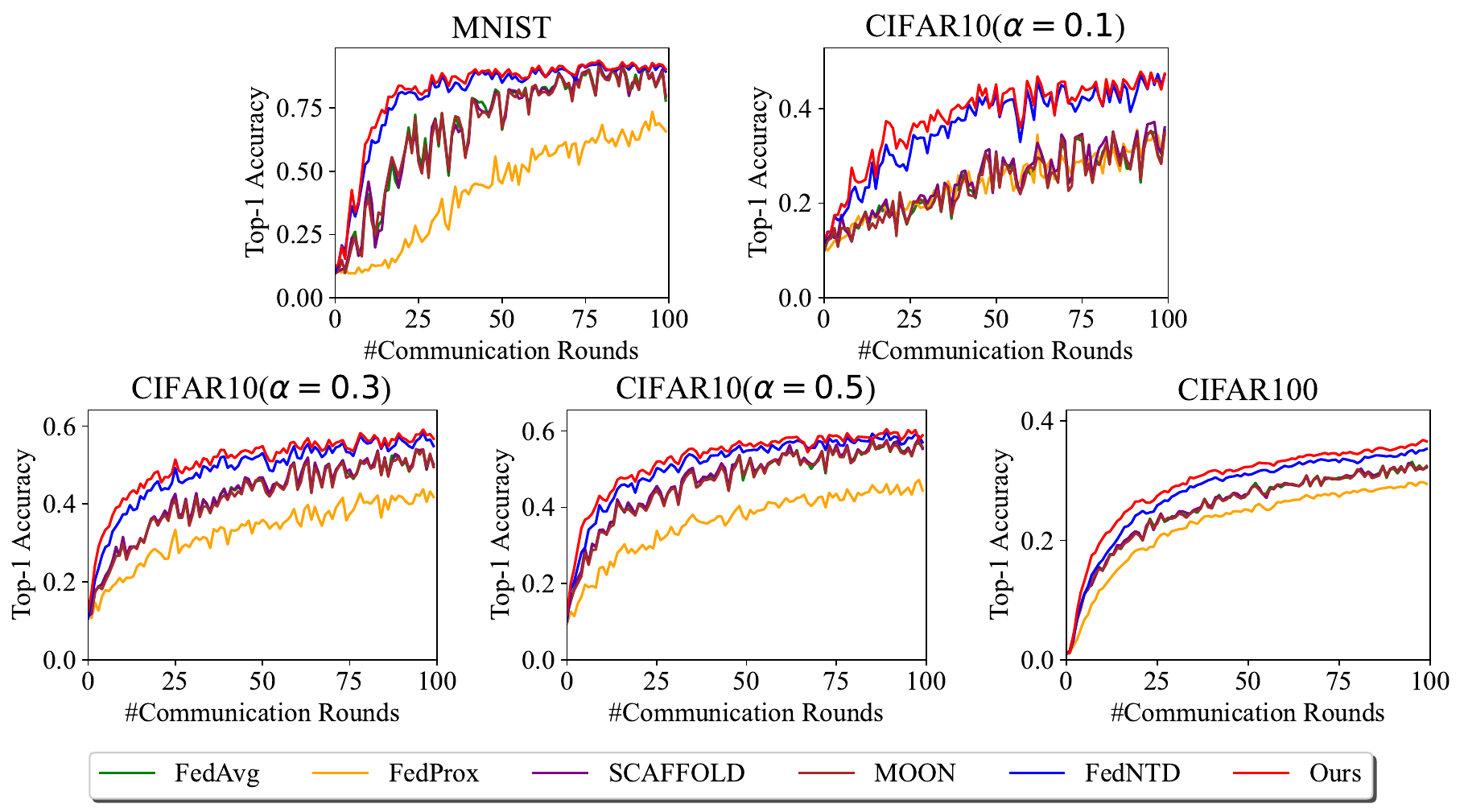}
\caption{Top-1 Accuracy}
\label{fig: top-1-acc}
\end{figure*}

The computational efficiency of federated learning methods is crucial for their practical deployment, especially considering the limited communication bandwidth and the computational resources of participating clients. Table \ref{tab:Com-Eff} and Fig. \ref{fig: top-1-acc} presents a comparative analysis of the communication efficiency across different federated learning approaches, including our proposed FedDistill method. This metric is defined by the number of communication rounds required by each method to reach the final accuracy benchmark established by FedAvg on various datasets.

Our observations from the results highlight several key points regarding the efficiency of FedDistill compared to established methods such as FedAvg, FedProx, SCAFFOLD, MOON, and FedNTD. (1) FedDistill significantly outperforms all baseline methods in terms of communication efficiency. It achieves the benchmark top-1 accuracy set by FedAvg with considerably fewer communication rounds across all datasets tested. For instance, on MNIST, FedDistill requires only 16 rounds compared to FedAvg's 100 rounds, showcasing a drastic reduction in communication needs. (2) The efficiency of FedDistill is particularly pronounced under various levels of non-IID data distribution ($\alpha$ values). As the non-IID level increases from 0.5 to 0.1, our method demonstrates an increasing advantage in communication efficiency, indicating its robustness and adaptability to different degrees of data heterogeneity. (3) When compared to more recent approaches like FedNTD, FedDistill not only achieves higher accuracy but also requires fewer rounds to do so. This efficiency becomes more significant with more pronounced non-IID distributions, highlighting the effectiveness of our method in handling the challenges posed by skewed data distributions. (4) The communication efficiency of FedDistill suggests that it can significantly reduce the communication cost associated with federated learning. This reduction is critical for real-world applications where bandwidth and communication costs are limiting factors, particularly in environments with constrained resources or high data privacy requirements.

\subsection{Ablation Study}

\begin{figure*}[!t]
\centering
\subfloat[$\alpha_t$]{\includegraphics[width=.5\textwidth]{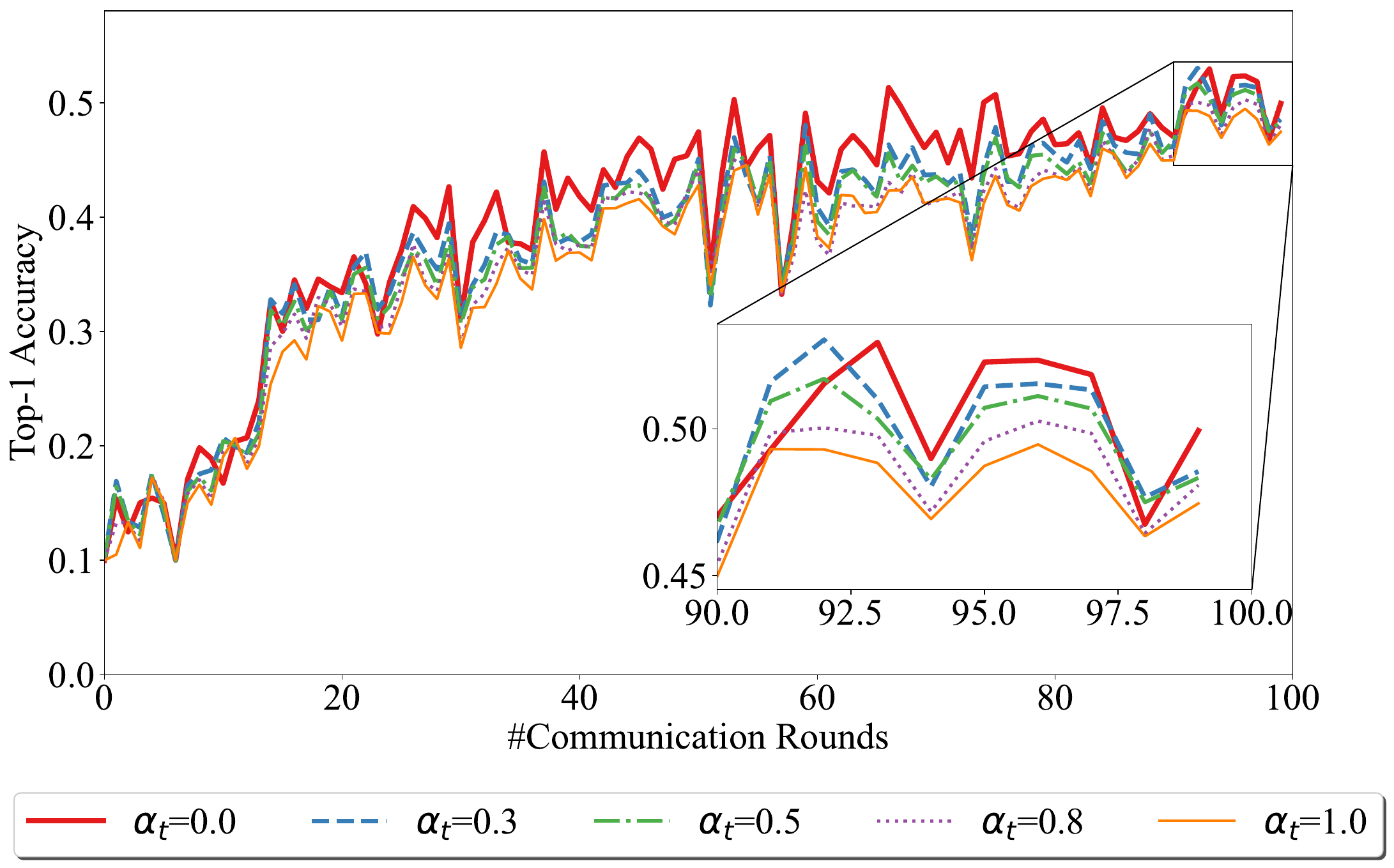}%
\label{fig:GD_t}}
\hfil
\subfloat[$\alpha_f$]{\includegraphics[width=.5\textwidth]{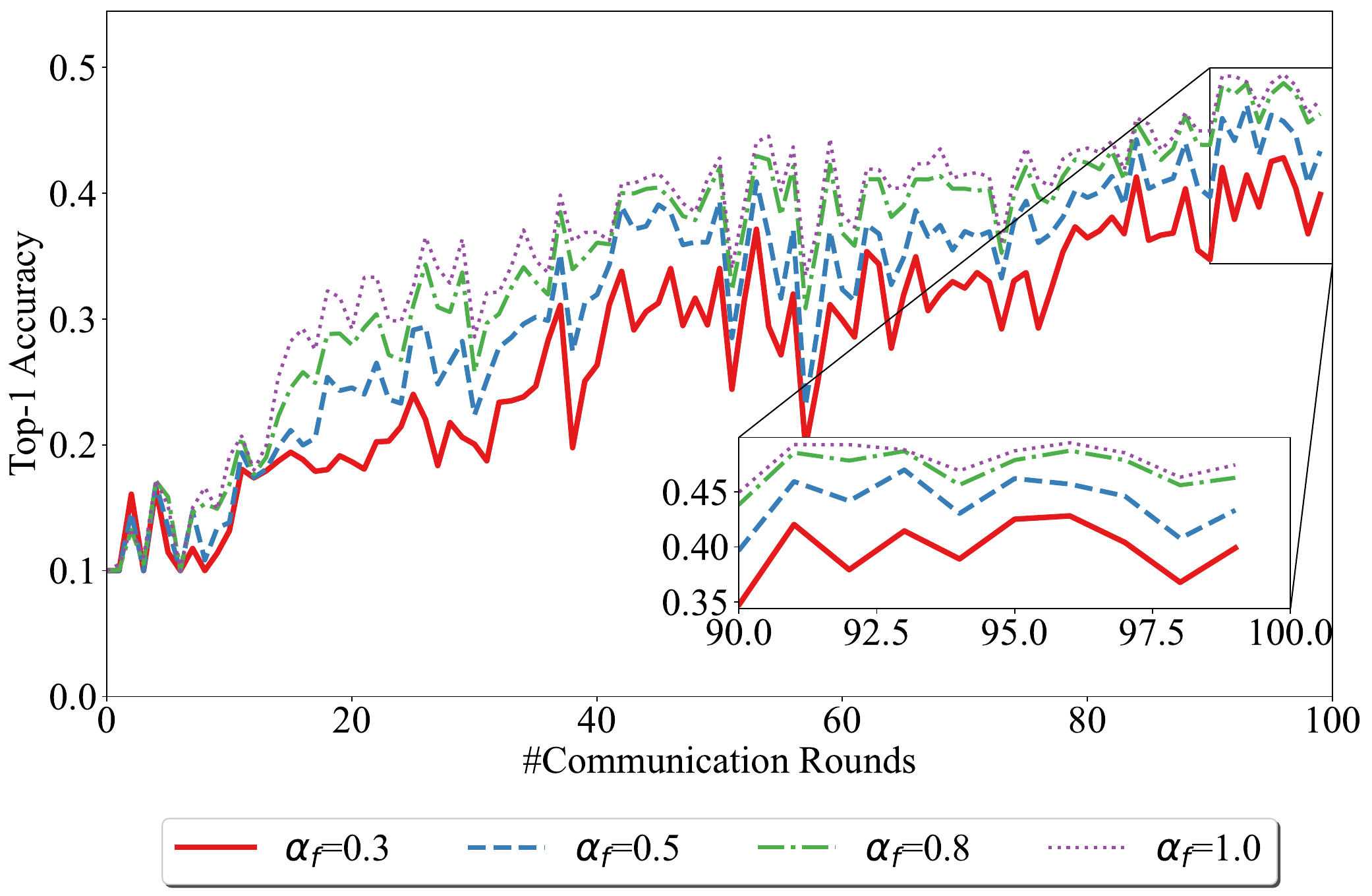}%
\label{fig:GD_f}}
\hfil
\subfloat[$\alpha_r$]{\includegraphics[width=.5\textwidth]{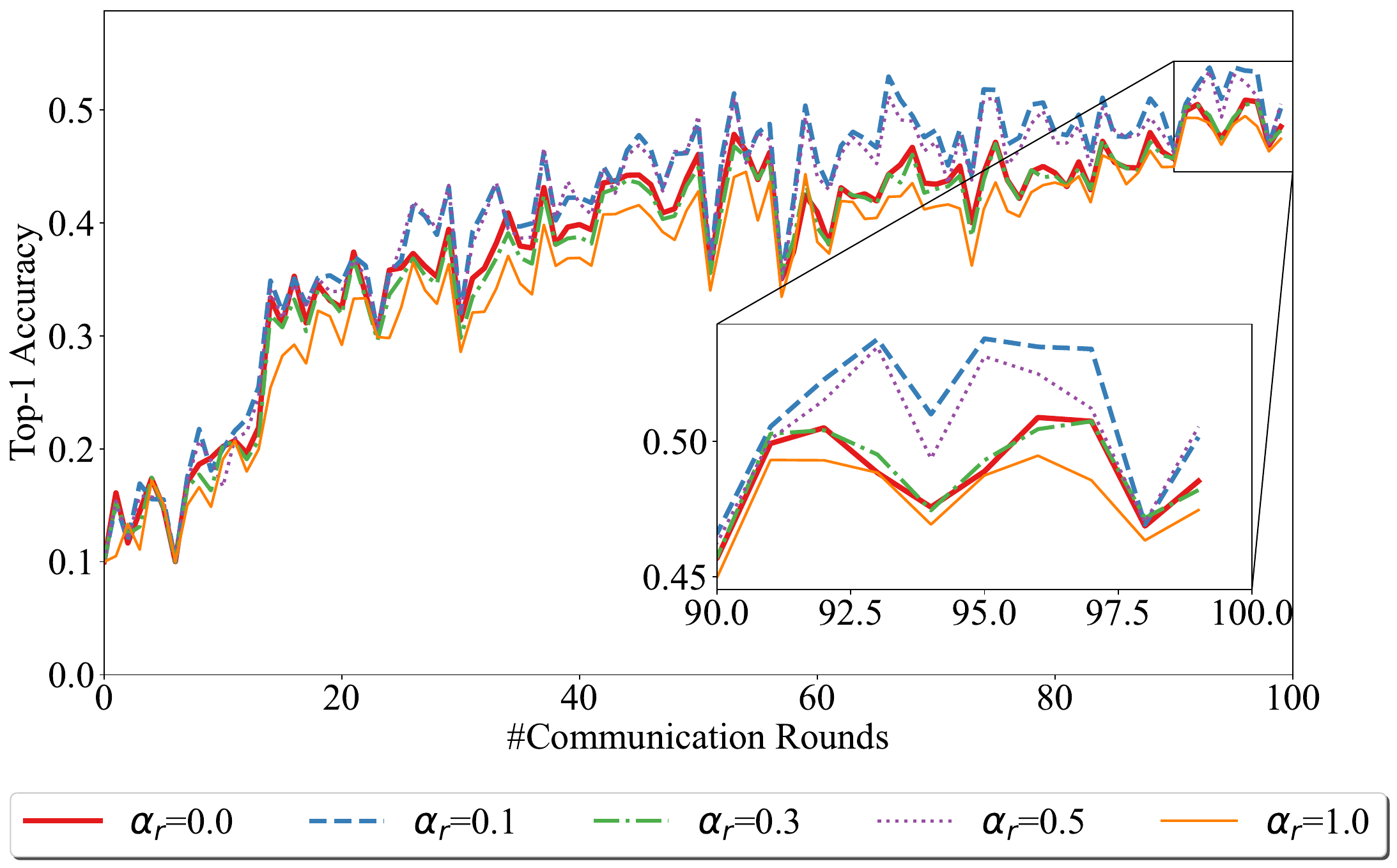}%
\label{fig:GD_r}}
\hfil
\subfloat[$\beta_{\text{L}}$]{\includegraphics[width=.5\textwidth]{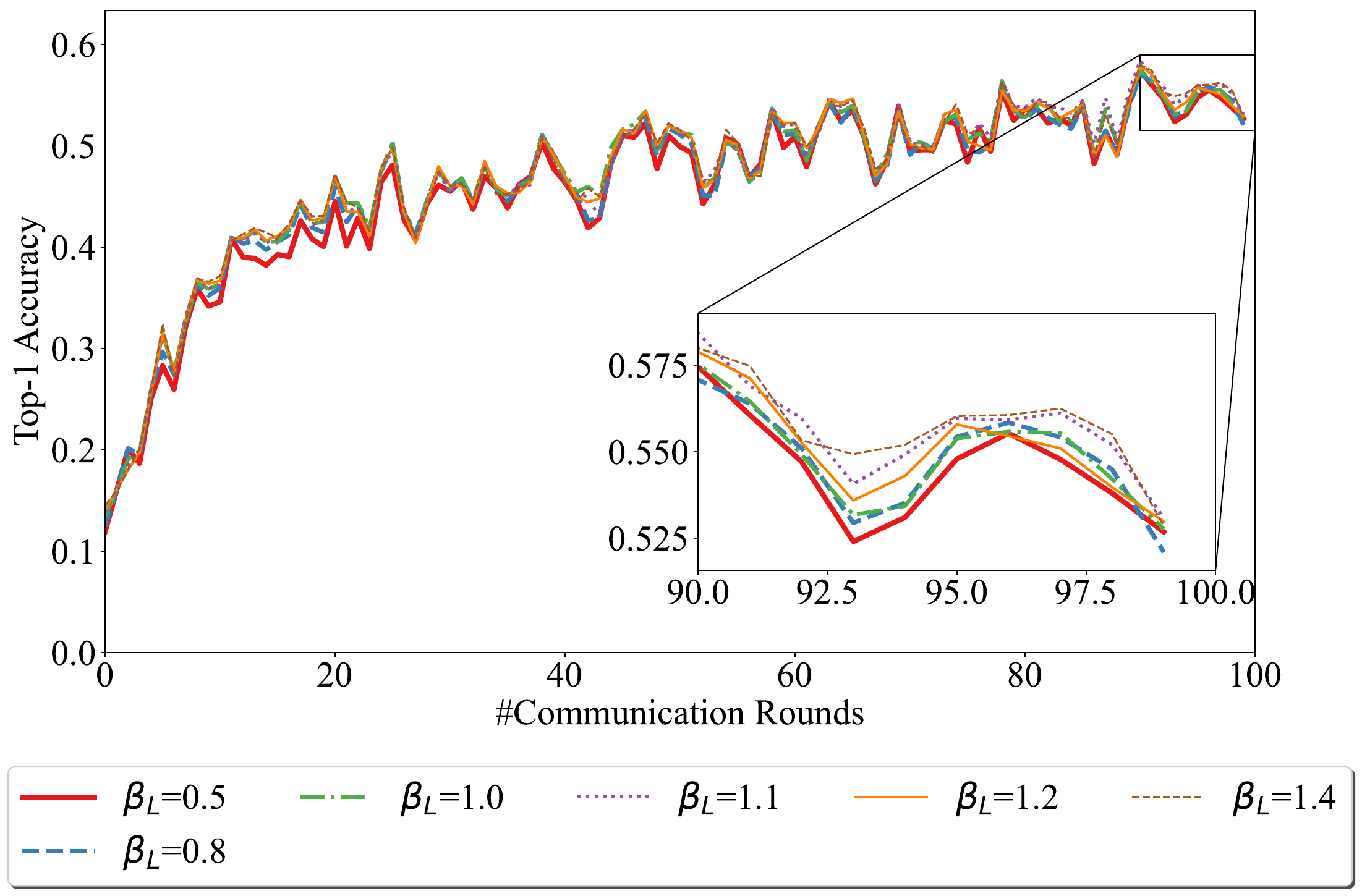}%
\label{fig:beta_L}}
\hfil
\subfloat[$\beta_{\text{FC}}$]{\includegraphics[width=.5\textwidth]{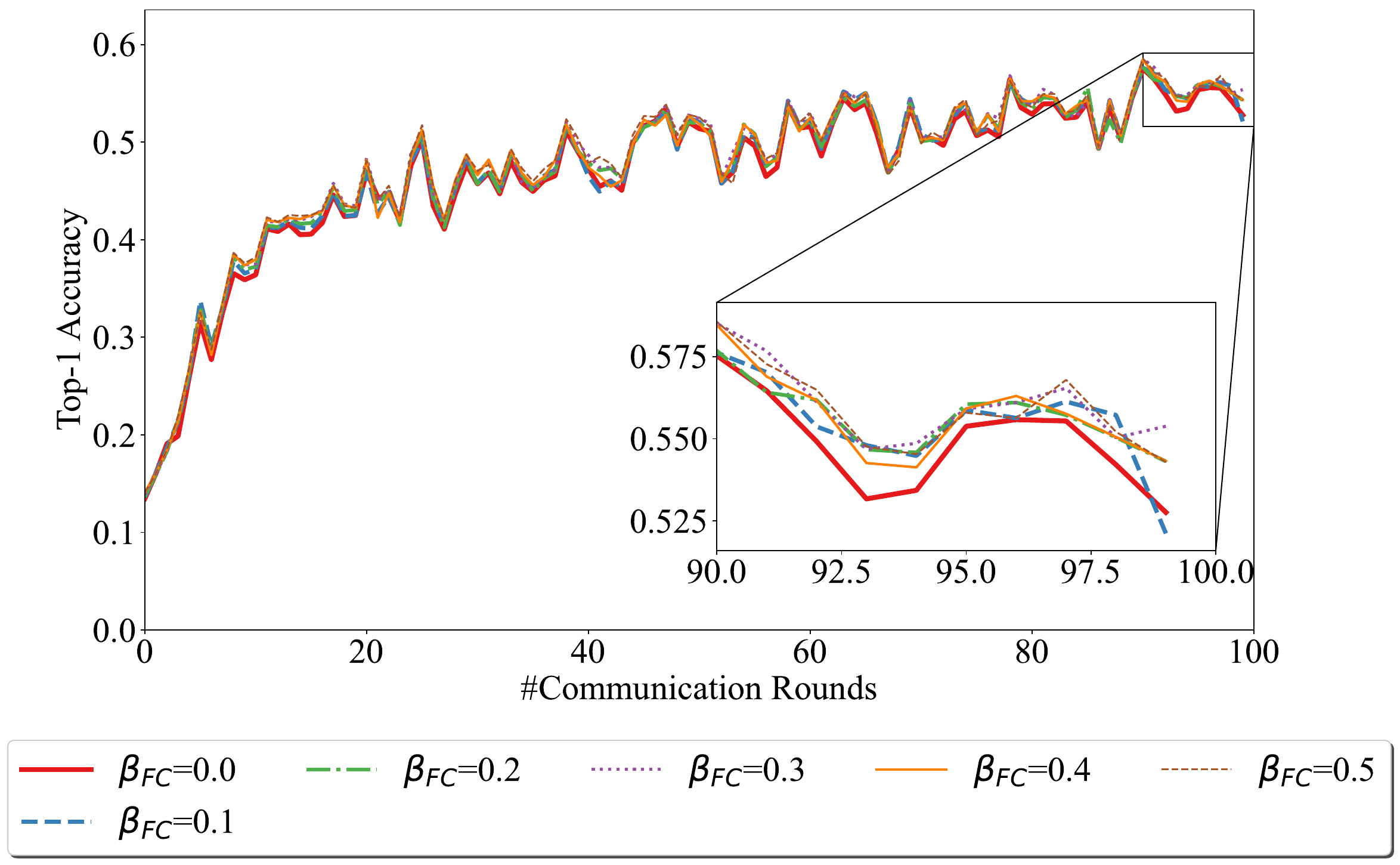}%
\label{fig:beta_FC}}
\hfil
\subfloat[$\beta_{\text{E}}$]{\includegraphics[width=.5\textwidth]{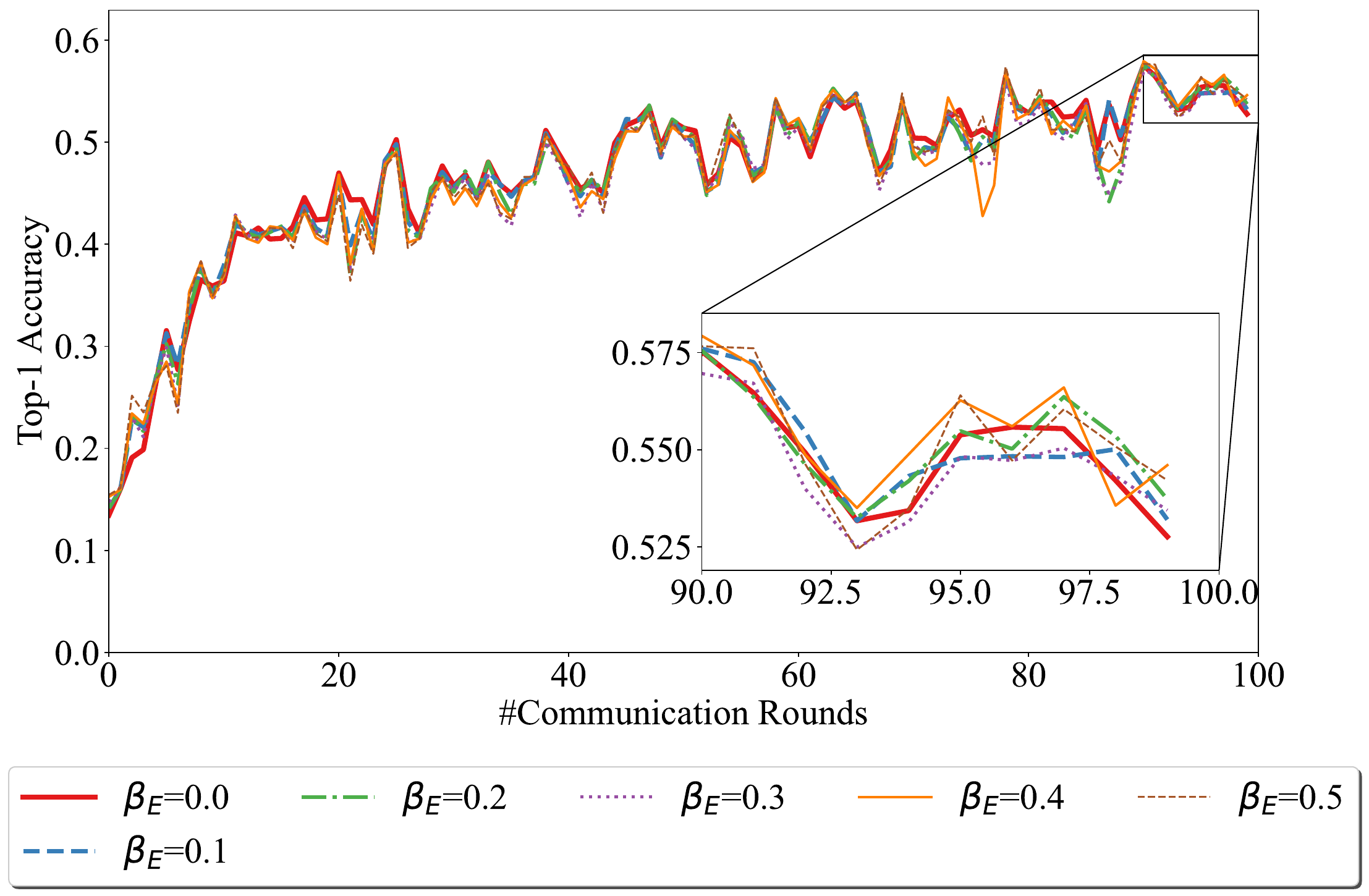}%
\label{fig:beta_E}}

\caption{Effects of hyperparameters on the performance of FedDistill.}
\label{fig:GD_hypers}
\end{figure*}

The ablation study, as presented in Table \ref{tab:ablation} and Fig. \ref{fig:GD_hypers}, systematically investigates the impact of each proposed loss component on the overall performance of our FL method. This analysis crucially dissects the role of each component to understand their contributions towards enhancing model generalization and addressing the challenge of data imbalance across clients.

\textbf{Impact of $\mathcal{L}$.} The most significant observation is the paramount importance of the $\mathcal{L}_E$ component. The observations from the results are as follows. (1) Removing $\mathcal{L}_E$ leads to a noticeable drop in top-1 accuracy (from 0.7082 to 0.6964), underscoring its essential role in enabling the local feature extractors to generalize effectively. This loss component, by facilitating the processing of local features through both biased and balanced (global) classifiers, ensures that the feature extractor is not solely tailored to the local data distribution but also retains the capability to learn and adapt based on the more generalized global model insights. (2) While the removal of $\mathcal{L}_{\text{FC}}$ also results in a decrease in performance (to 0.7050), the impact is less pronounced compared to $\mathcal{L}_E$. This observation suggests that while de-biasing the local classifier is beneficial for enhancing model performance, the more critical factor for generalization lies in the feature extraction process. (3) From Fig. \ref{fig:beta_L}, we observe that although top-1 accuracy of last epoch is relatively robust for different $\beta_{\text{L}}$ the average top-1 accuracy of last 10 epochs is increased with higher $\beta_{\text{L}}$. This observation demonstrates that $\mathcal{L}_{\text{L}}$ also contributes to the training stability. (4) The study clearly demonstrates that both $\mathcal{L}_E$ and $\mathcal{L}_{\text{FC}}$ are integral to achieving the method's high performance, with $\mathcal{L}_E$ identified as a more influential factor. This aligns with the earlier discussion on the imbalanced local data's effect on the model, particularly highlighting the classifier's vulnerability to such imbalances.

This study reveals the indispensable role of $\mathcal{L}_E$ in ensuring the local feature extractor's generalization capability by leveraging a dual classifier approach. It highlights the necessity of addressing both feature extraction and classification processes to combat the challenges posed by imbalanced local data in Federated Learning. 

\begin{table}[!t]
\caption{Ablation Study (Sample ratio=100\%)}\label{tab:ablation}
\centering
\begin{tabular}{cc}
\hline
\multirow{2}*{Methods} & CIFAR10  \\
~ & $\alpha=0.5$   \\
\hline
\hline
Ours & 0.7082 \\
\hline
w/o $\mathcal{L}_{\text{E}}$ & 0.6964  \\
w/o $\mathcal{L}_{\text{FC}}$ & 0.7050  \\
\end{tabular}
\end{table}

\begin{figure}[!t]
\captionof{table}{Demonstration of group distillation loss impact on CIFAR10 accuracy, showcasing the method's effectiveness in improving learning outcomes for few-sample classes. (Sample ration=100\%, Non-iid level=0.5)}
\centering
\begin{tabular}{c|ccc|cc}
\hline
Dataset & TC-KD & FC-KD & RC-KD & Top-1 & $\Delta$ \\
\hline
\multirow{7}*{CIFAR10} &  \multicolumn{3}{c|}{Baseline} & 67.7  \\
\cline{2-6}
~ & \checkmark & & &  67.34 & -0.36  \\
~ &  & \checkmark & &  70.03 & +2.33 \\
~ &  &  & \checkmark &  66.8 & -0.9 \\
~ & \checkmark & \checkmark &  &  69.92 & +2.22  \\
~ & \checkmark &  & \checkmark &  66.62 & -1.08 \\
~ &  & \checkmark & \checkmark &  69.14 & +1.44 \\
\hline
\end{tabular}
\label{tab:GD}
\end{figure}

\textbf{Impact of group distillation loss.} The results from the demonstration of group distillation loss's impact on CIFAR10 accuracy highlight the significant role of each component within the group distillation process and its effectiveness in addressing few-sample class challenges (see Table. \ref{tab:GD}). The baseline performance on CIFAR10 is set at a Top-1 accuracy of 67.7\%, serving as a benchmark for comparing the influence of incorporating True-Class KD (TC-KD), Few-sample Class KD (FC-KD), and Rich-sample Class KD (RC-KD) components individually and in combination.

The observations from the results are as follows: (1) The inclusion of FC-KD alone leads to a substantial increase in accuracy (+2.33\%), indicating its critical role in enhancing learning outcomes for few-sample classes. This suggests that focusing on underrepresented classes through targeted distillation significantly contributes to overall model performance. (2) When TC-KD and FC-KD are combined, there's a notable improvement (+2.22\%) over the baseline, reaffirming the importance of addressing both true-class and few-sample classes simultaneously to achieve a balanced and effective learning strategy. (3) Interestingly, implementing RC-KD alone or in combination with TC-KD results in a decrease in accuracy, suggesting that mere emphasis on well-represented classes may detract from the model's ability to generalize across the broader class spectrum. (4) The combination of FC-KD with RC-KD without TC-KD also results in an increase in accuracy (+1.44\%), highlighting the nuanced interplay between focusing on few-sample and rich-sample classes. This combination potentially mitigates the overemphasis on well-represented classes while still benefiting from the targeted approach towards underrepresented classes.

More specifically, Fig. \ref{fig:GD_t} - \ref{fig:GD_r} demonstrates the influence of $\alpha_t$, $\alpha_r$, and $\alpha_f$ on top-1 accuracy on CIFAR10 with non-iid level as 0.3. The figure shows that (1) FC-KD ($\alpha_f$) plays the most important part of our Group Distillation loss: from $0.3$ to $0.5$ and from $0.5$ to $0.8$ the top-1 accuracy was significantly improved, but from $0.8$ to $1.0$ the improvement was not as significant as it was. (2) RC-KD also contribute to our Group Distillation loss, but a high(1.0) or low(0.0) $\alpha_r$ both result in a decrease on top-1 accuracy. It demonstrate that although RC-KD contribute to the performance, it should be suppress to yield a better top-1 accuracy. (3) TC-KD shows a negative impact on the top-1 accuracy: from $0.0$ to $1.0$ the increase of $\alpha_t$ correspond to the decrease of performances.

Generally speaking, the result of Fig. \ref{fig:GD_hypers} is aligned with Table. \ref{tab:GD}, but in a more detailed form. These findings highlight the complexity of managing class imbalances in federated learning environments. The significant improvement observed with the implementation of FC-KD alone and in combination with other components underscores the value of specifically addressing few-sample classes to enhance overall model performance.




\section{Conclusion}

In this study, we identified the primary cause of local model forgetting in non-IID FL as the inadequate positive reinforcement for classes that are underrepresented in local datasets. To address this issue, we introduced the concept of Group Distillation, designed to retain the global model's knowledge for these few-sample classes within local clients. Additionally, we proposed the decomposition of the global model into a balanced feature extractor and classifier, aiming to counteract the tendency of local models to develop biases due to imbalanced data distributions.

Our evaluation shows that our method not only surpasses a range of baseline approaches in various datasets but also demonstrates accelerated convergence. These findings highlight our method's capacity to enhance the accuracy and generalization of FL models.

While our method is fast-convergent, and generic, we acknowledge the complexity of training induced by the additional hyper-parameters. To address this, in the future, we will focus on introducing self-adaptive learning approach.




\begin{thebibliography}{1}

\bibitem{ref1}  Luo M, Chen F, Hu D, et al. No fear of heterogeneity: Classifier calibration for federated learning with non-iid data[J].  Advances in Neural Information Processing Systems, 2021, 34: 5972-5984.

\bibitem{ref2}  McMahan B, Moore E, Ramage D, et al. Communication-efficient learning of deep networks from decentralized data[C]//Artificial intelligence and statistics. PMLR, 2017: 1273-1282.

\bibitem{ref3}  Lee G, Jeong M, Shin Y, et al. Preservation of the Global Knowledge by Not-True Distillation in Federated Learning[J]. arXiv preprint arXiv:2106.03097, 2021.

\bibitem{ref4}  Xu C, Hong Z, Huang M, et al. Acceleration of federated learning with alleviated forgetting in local training[J]. arXiv preprint arXiv:2203.02645, 2022.

\bibitem{ref5}  Zhao Y, Li M, Lai L, et al. Federated learning with non-iid data[J]. arXiv preprint arXiv:1806.00582, 2018.

\bibitem{ref6}  Hong-You Chen and Wei-Lun Chao. Fedbe: Making bayesian model ensemble applicable to federated learning. arXiv preprint arXiv:2009.01974, 2020.

\bibitem{ref7}  Tao Lin, Lingjing Kong, Sebastian U Stich, and Martin Jaggi. Ensemble distillation for robust model fusion in federated learning. arXiv preprint arXiv:2006.07242, 2020.

\bibitem{ref8}  Hongyi Wang, Mikhail Yurochkin, Yuekai Sun, Dimitris Papailiopoulos, and Yasaman Khazaeni. Federated learning with matched averaging. arXiv preprint arXiv:2002.06440, 2020.

\bibitem{ref9}  Lin Zhang, Li Shen, Liang Ding, Dacheng Tao, and Ling-Yu Duan. Fine-tuning global model via data-free knowledge distillation for non-iid federated learning. arXiv preprint arXiv:2203.09249, 2022.

\bibitem{ref10}  M. Mendieta, T. Yang, P. Wang, M. Lee, Z. Ding and C. Chen, "Local Learning Matters: Rethinking Data Heterogeneity in Federated Learning," 2022 IEEE/CVF Conference on Computer Vision and Pattern Recognition (CVPR), New Orleans, LA, USA, 2022, pp. 8387-8396, doi: 10.1109/CVPR52688.2022.00821.

\bibitem{ref11}  L. Zhang, Y. Luo, Y. Bai, B. Du and L. -Y. Duan, "Federated Learning for Non-IID Data via Unified Feature Learning and Optimization Objective Alignment," 2021 IEEE/CVF International Conference on Computer Vision (ICCV), Montreal, QC, Canada, 2021, pp. 4400-4408, doi: 10.1109/ICCV48922.2021.00438.

\bibitem{ref12}  Karimireddy S P, Kale S, Mohri M, et al. Scaffold: Stochastic controlled averaging for federated learning[C]//International Conference on Machine Learning. PMLR, 2020: 5132-5143.

\bibitem{MOON}  Li Q, He B, Song D. Model-contrastive federated learning[C]//Proceedings of the IEEE/CVF Conference on Computer Vision and Pattern Recognition. 2021: 10713-10722.

\bibitem{ref14}  Neta Shoham, Tomer Avidor, Aviv Keren, Nadav Israel, Daniel Benditkis, Liron Mor-Yosef, and Itai Zeitak. Overcoming forgetting in federated learning on non-iid data. arXiv preprint arXiv:1910.07796, 2019.


\bibitem{ref16}  Legate G, Caccia L, Belilovsky E. Re-Weighted Softmax Cross-Entropy to Control Forgetting in Federated Learning[J]. arXiv preprint arXiv:2304.05260, 2023.

\bibitem{ref17}  Li Z, Shao J, Mao Y, et al. Federated learning with gan-based data synthesis for non-iid clients[C]//Trustworthy Federated Learning: First International Workshop, FL 2022, Held in Conjunction with IJCAI 2022, Vienna, Austria, July 23, 2022, Revised Selected Papers. Cham: Springer International Publishing, 2023: 17-32.

\bibitem{ref18}  Zhao Y, Li M, Lai L, et al. Federated learning with non-iid data[J]. arXiv preprint arXiv:1806.00582, 2018.


\bibitem{ref20}  Zhao B, Cui Q, Song R, et al. Decoupled knowledge distillation[C]//Proceedings of the IEEE/CVF Conference on computer vision and pattern recognition. 2022: 11953-11962.

\bibitem{ref21}  Fang G, Song J, Shen C, et al. Data-free adversarial distillation[J]. arXiv preprint arXiv:1912.11006, 2019.

\bibitem{ref22}  Q. Li, Y. Diao, Q. Chen and B. He, "Federated Learning on Non-IID Data Silos: An Experimental Study," 2022 IEEE 38th International Conference on Data Engineering (ICDE), Kuala Lumpur, Malaysia, 2022, pp. 965-978, doi: 10.1109/ICDE53745.2022.00077.

\bibitem{ref23}  Tian Li, Anit Kumar Sahu, Manzil Zaheer, Maziar Sanjabi, Ameet Talwalkar, and Virginia Smith. Federated optimization in heterogeneous networks. Proceedings of Machine Learning and Systems (MLSys), 2020.

\bibitem{ref24}  Yin Cui, Menglin Jia, Tsung-Yi Lin, Yang Song, and Serge Belongie. Class-balanced loss based on effective number of samples. In Proceedings of the IEEE/CVF Conference on Computer Vision and Pattern Recognition, pages 92689277, 2019.

\bibitem{reweight_softmax}  Legate G, Caccia L, Belilovsky E. Re-weighted softmax cross-entropy to control forgetting in federated learning[J]. arXiv preprint arXiv:2304.05260, 2023.

\bibitem{KD}  Hinton G, Vinyals O, Dean J. Distilling the knowledge in a neural network[J]. arXiv preprint arXiv:1503.02531, 2015.

\bibitem{SD}  Pham M, Cho M, Joshi A, et al. Revisiting self-distillation[J]. arXiv preprint arXiv:2206.08491, 2022.

\bibitem{ref28}  X. Shang, Y. Lu, Y. -M. Cheung and H. Wang, "FEDIC: Federated Learning on Non-IID and Long-Tailed Data via Calibrated Distillation," 2022 IEEE International Conference on Multimedia and Expo (ICME), Taipei, Taiwan, 2022, pp. 1-6, doi: 10.1109/ICME52920.2022.9860009

\bibitem{FedDistill_2022}  Seo H, Park J, Oh S, et al. 16 federated knowledge distillation[J]. Machine Learning and Wireless Communications, 2022: 457.

\bibitem{fm}  Arslan C, Puneet D, Thalaiyasingam A, et al. Riemannian walk for incremental learning: Understanding forgetting and intransigence. In Proceedings of the European Conference on Computer Vision (ECCV), pages 532–547, 2018.

\bibitem{long-tail survey}  Zhang Y, Kang B, Hooi B, et al. Deep long-tailed learning: A survey[J]. IEEE Transactions on Pattern Analysis and Machine Intelligence, 2023.




\bibitem{imbalanced_survey}  Zhang J, Li C, Qi J, et al. A Survey on Class Imbalance in Federated Learning[J]. arXiv preprint arXiv:2303.11673, 2023.

\bibitem{FedAux} Sattler F, Korjakow T, Rischke R, et al. Fedaux: Leveraging unlabeled auxiliary data in federated learning[J]. IEEE Transactions on Neural Networks and Learning Systems, 2021, 34(9): 5531-5543.

\bibitem{medical2} Tang T, Han Z, Cai Z, et al. Personalized Federated Graph Learning on Non-IID Electronic Health Records[J]. IEEE Transactions on Neural Networks and Learning Systems, 2024.

\bibitem{recommendation2} Sun Z, Xu Y, Liu Y, et al. A Survey on Federated Recommendation Systems[J]. IEEE Transactions on Neural Networks and Learning Systems, 2024.

\bibitem{edge2} Jiang Y, Wang S, Valls V, et al. Model pruning enables efficient federated learning on edge devices[J]. IEEE Transactions on Neural Networks and Learning Systems, 2022.

\end{thebibliography}
\end{document}